\documentclass[10pt,twocolumn,letterpaper]{article}

\usepackage[pagenumbers]{comprecap}  

\usepackage{capt-of}
\usepackage{xspace}
\usepackage{xcolor}
\usepackage{enumitem}
\usepackage{amsmath,amsthm,amssymb,amsfonts,dsfont,pifont,bm,bbm,mathrsfs,mathtools,nicefrac,extarrows,relsize}
\usepackage{algorithm,algpseudocode,listings}
\usepackage{booktabs,multirow,adjustbox,diagbox,threeparttable,tabularray,setspace}
\usepackage[misc]{ifsym}
\usepackage{lipsum}
\usepackage{tablefootnote} 

\definecolor{comprecapblue}{rgb}{0.21,0.49,0.74}
\usepackage[pagebackref,breaklinks,colorlinks,citecolor=comprecapblue,bookmarks=false]{hyperref}
\usepackage{wrapfig}
\usepackage[capitalize]{cleveref}  
\newcommand\blfootnote[1]{%
  \begingroup
  \renewcommand\thefootnote{}\footnote{#1}%
  \addtocounter{footnote}{-1}%
  \endgroup
}

\crefname{section}{Sec.}{Secs.}
\Crefname{section}{Section}{Sections}
\crefname{appendix}{App.}{Apps.}
\Crefname{appendix}{Appendix}{Appendices}
\crefname{table}{Tab.}{Tabs.}
\Crefname{table}{Table}{Tables}
\crefname{figure}{Fig.}{Figs.}
\Crefname{figure}{Figure}{Figures}
\crefname{equation}{Eq.}{Eqs.}
\Crefname{equation}{Equation}{Equations}
\crefname{theorem}{Thm.}{Thms.}
\Crefname{theorem}{Theorem}{Theorems}
\crefname{lemma}{Lem.}{Lems.}
\Crefname{lemma}{Lemma}{Lemmas}
\crefname{remark}{Rem.}{Rems.}
\Crefname{remark}{Remark}{Remarks}
\crefname{corollary}{Cor.}{Cors.}
\Crefname{corollary}{Corollary}{Corollaries}
\crefname{algorithm}{Alg.}{Algs.}
\Crefname{algorithm}{Algorithm}{Algorithms}
\definecolor{cellred}{RGB}{213, 123, 101}
\definecolor{cellgreen}{RGB}{0, 205, 0}
\definecolor{cellblue}{RGB}{54, 125, 189}
\definecolor{codegreen}{rgb}{0,0.6,0}
\definecolor{codegray}{rgb}{0.5,0.5,0.5}
\definecolor{codepurple}{rgb}{0.58,0,0.82}
\definecolor{backcolour}{rgb}{1.0,1.0,1.0}
\lstdefinestyle{mystyle}{
    backgroundcolor=\color{backcolour},
    commentstyle=\color{codegreen},
    keywordstyle=\color{magenta},
    numberstyle=\tiny\color{codegray},
    stringstyle=\color{codepurple},
    basicstyle=\ttfamily\scriptsize,
    breakatwhitespace=false,
    breaklines=true,
    captionpos=b,
    keepspaces=true,
    numbers=left,
    numbersep=5pt,
    showspaces=false,
    showstringspaces=false,
    showtabs=false,
    tabsize=2
}
\newcommand{\tocite}[1]{{\color{red} [TO CITE]}}
\newcommand{\methodname}{CompreCap}
\newcommand{\method}{\textit{\methodname}\xspace}
\newcommand{\supp}{\textit{Supplementary Material}\xspace}

\usepackage{graphicx} 
\usepackage{subcaption}
\usepackage{orcidlink}
\usepackage{url}
\usepackage{wrapfig}
\usepackage{color, colortbl}
\usepackage{physics}
\usepackage{multirow}
\usepackage{xcolor}
\usepackage{footnote}
\usepackage{float}
\usepackage{lipsum}
\usepackage{multicol}
\usepackage{siunitx}
\usepackage{tabularx,booktabs}
\usepackage{etoolbox}
\usepackage{threeparttable}
\usepackage{amsmath,amssymb,caption,multirow,overpic,textpos}
\usepackage{adjustbox,diagbox,tabularray,tcolorbox}
\usepackage{dsfont,pifont,bm,bbm,mathrsfs,mathtools,nicefrac}
\usepackage{algorithm,algpseudocode,listings,array}
\usepackage{titletoc}

\newcolumntype{x}[1]{>{\centering\arraybackslash}p{#1pt}}
\newcolumntype{y}[1]{>{\raggedright\arraybackslash}p{#1pt}}
\newcolumntype{z}[1]{>{\raggedleft\arraybackslash}p{#1pt}}
\newcommand{\red}[1]{\textcolor{red}{#1}}

\def\mA{{\mathcal A}}

\def\mS{{\mathcal S}}

\DeclareMathAlphabet\mathbfcal{OMS}{cmsy}{b}{n}

\def\0{{\bf 0}}
\def\1{{\bf 1}}

\def\eg{\emph{e.g.}} 
\def\ie{\emph{i.e.}}

\usepackage{amsmath}

\newlength\savewidth

\definecolor{baselinecolor}{gray}{.9}

\newcommand{\std}[1]{\tiny{$\pm \emph{#1}$}}

\usepackage{amsmath,amsfonts,bm}

\def\eqref#1{equation~\ref{#1}}

\def\1{\bm{1}}

\def\mA{{\bm{A}}}

\def\mS{{\bm{S}}}

\DeclareMathAlphabet{\mathsfit}{\encodingdefault}{\sfdefault}{m}{sl}
\SetMathAlphabet{\mathsfit}{bold}{\encodingdefault}{\sfdefault}{bx}{n}

\def\sR{{\mathbb{R}}}

\def\emA{{A}}

\def\emM{{M}}

\def\emS{{S}}

\title{Benchmarking Large Vision-Language Models via Directed Scene Graph\\for Comprehensive Image Captioning}

\author{
Fan Lu$^1$$^*$,  Wei Wu$^1$$^*$, Kecheng Zheng$^2$\textsuperscript{\Letter}, Shuailei Ma$^3$, Biao Gong$^2$, \\ 
Jiawei Liu$^1$, Wei Zhai$^1$\textsuperscript{\Letter}, Yang Cao$^1$, Yujun Shen$^2$, Zheng-Jun Zha$^1$\\
$^1$ MoE Key Laboratory of Brain-inspired Intelligent Perception and Cognition, \\ University of Science and Technology of China \quad
$^2$ Ant Group  \quad
$^3$ Northeastern University, China
}

\begin{document}
\twocolumn[{%
\renewcommand\twocolumn[1][]{#1}%
\maketitle
\begin{center}
\vspace{-10pt}
\includegraphics[width=0.99\textwidth]{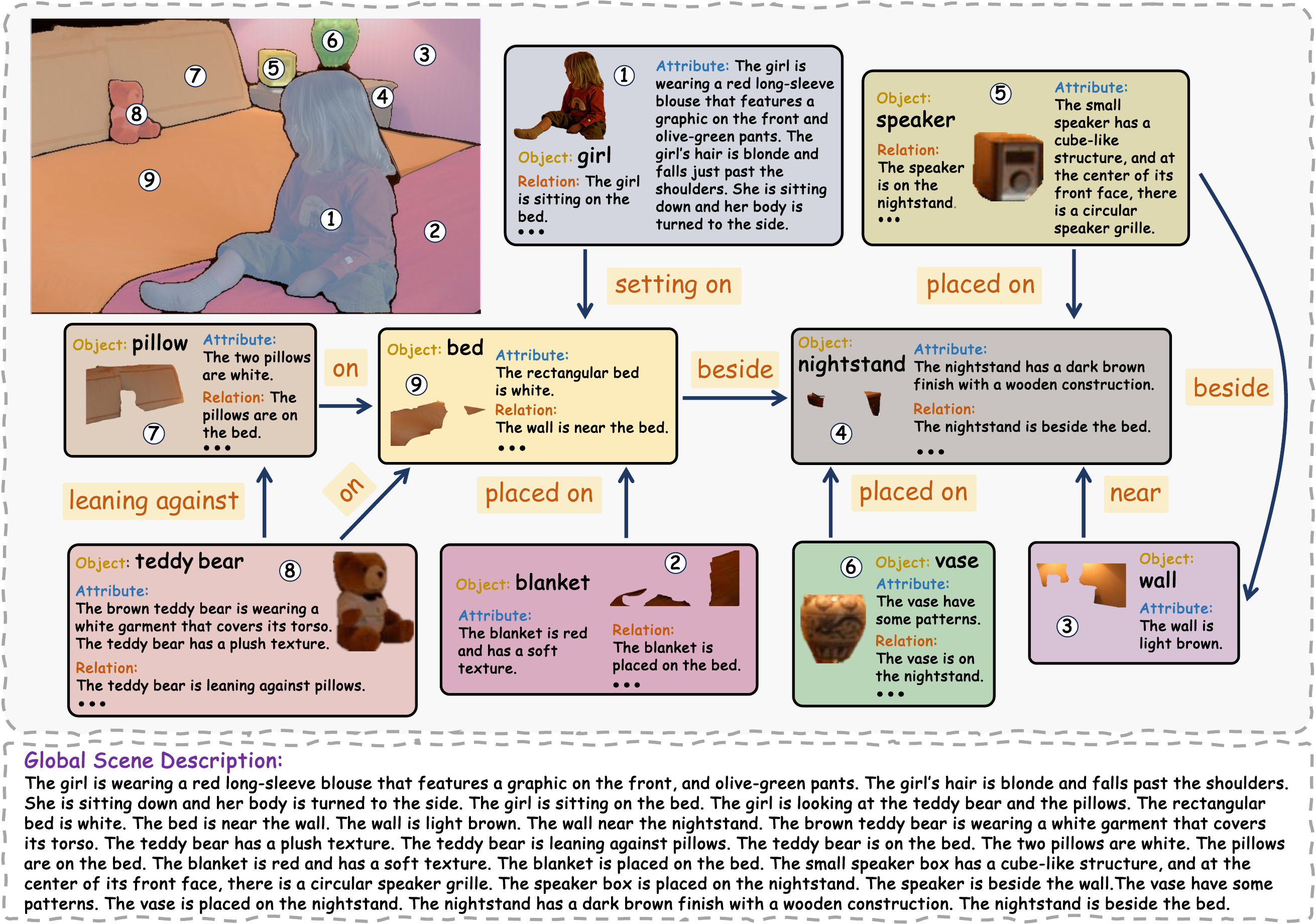}
\vspace{-1.5mm}
\captionof{figure}{
\textbf{
    Data sample of \method dataset}. 
    Objects in the image are manually annotated with segmentation maps, semantic labels, their bound attribute descriptions and the relations with other objects, composing a directed scene graph structure.
    The global scene description is organized by these annotations.
    }
\label{fig:data_struct}
\end{center}%
}]

\blfootnote{\small *~: Equal contribution. \Letter~: Corresponding author.}
\begin{abstract}

Generating detailed captions comprehending text-rich visual content in images has received growing attention for Large Vision-Language Models (LVLMs).
However, few studies have developed benchmarks specifically tailored for detailed captions to measure their accuracy and comprehensiveness.
%
In this paper, we introduce a detailed caption benchmark, termed as \method, to evaluate the visual context from a directed scene graph view.
%
%
Concretely, we first manually segment the image into semantically meaningful regions (i.e., semantic segmentation mask) according to common-object vocabulary, while also distinguishing attributes of objects within all those regions.
Then directional relation labels of these objects are annotated to compose a directed scene graph that can well encode rich compositional information of the image.
Based on our directed scene graph, we develop a pipeline to assess the generated detailed captions from LVLMs on multiple levels, including the object-level coverage, the accuracy of attribute descriptions, the score of key relationships, {etc}.
Experimental results on the \method dataset confirm that our evaluation method aligns closely with human evaluation scores across LVLMs.
We have released the code and the dataset \href{https://github.com/LuFan31/CompreCap}{here} to support the community.

\begin{table*}[h]
    \captionsetup{skip=3pt}
    \caption{Comparison of \method with prior caption evaluation dataset in annotation components. All datasets are annotated \textbf{by human}.}
    \label{tab:data_composition}
    \centering
    \scriptsize
    \SetTblrInner{rowsep=0.5pt}      
    \SetTblrInner{colsep=9.0pt}      
    \resizebox{\linewidth}{!}{
    \begin{tblr}{
        cells={halign=c,valign=m},   
        hline{1,2,9}={1.0pt},         
        hline{8}={},         
    }
        Dataset  & {Averaged\\ Text Length } & Object & {Segmentation\\ Map} & Attribute & Relation & Q/A & Answer Type \\
        
         MSCOCO~\cite{lin2014microsoft}  & 10 & \checkmark & \checkmark & - & - & - & - \\
         NoCaps~\cite{agrawal2019nocaps}  & 11 & \checkmark & - & - & - & -& - \\
         POPE~\cite{li2023evaluating}  & -  & \checkmark & - & -&- &\checkmark & Y/N \\
         FGHE~\cite{wang2024mitigating} & -  & \checkmark & - & -&- &\checkmark & Y/N  \\
         Q-Bench~\cite{wu2023q}  & 58 & - & -& - &- &\checkmark & Y/N and A/B/C \\
         DetailCaps-100~\cite{detailcaption} & 177 & - & -& - &- & - & - \\
         \method & 172 & \checkmark & \checkmark  & \checkmark& \checkmark& \checkmark & A/B/C\\
        \end{tblr}
        }
\vspace{-5pt}
\end{table*}

\end{abstract}
     
\section{Introduction}\label{sec:intro}

\begin{figure*}[t]
    \centering
    \includegraphics[width=0.99\textwidth]{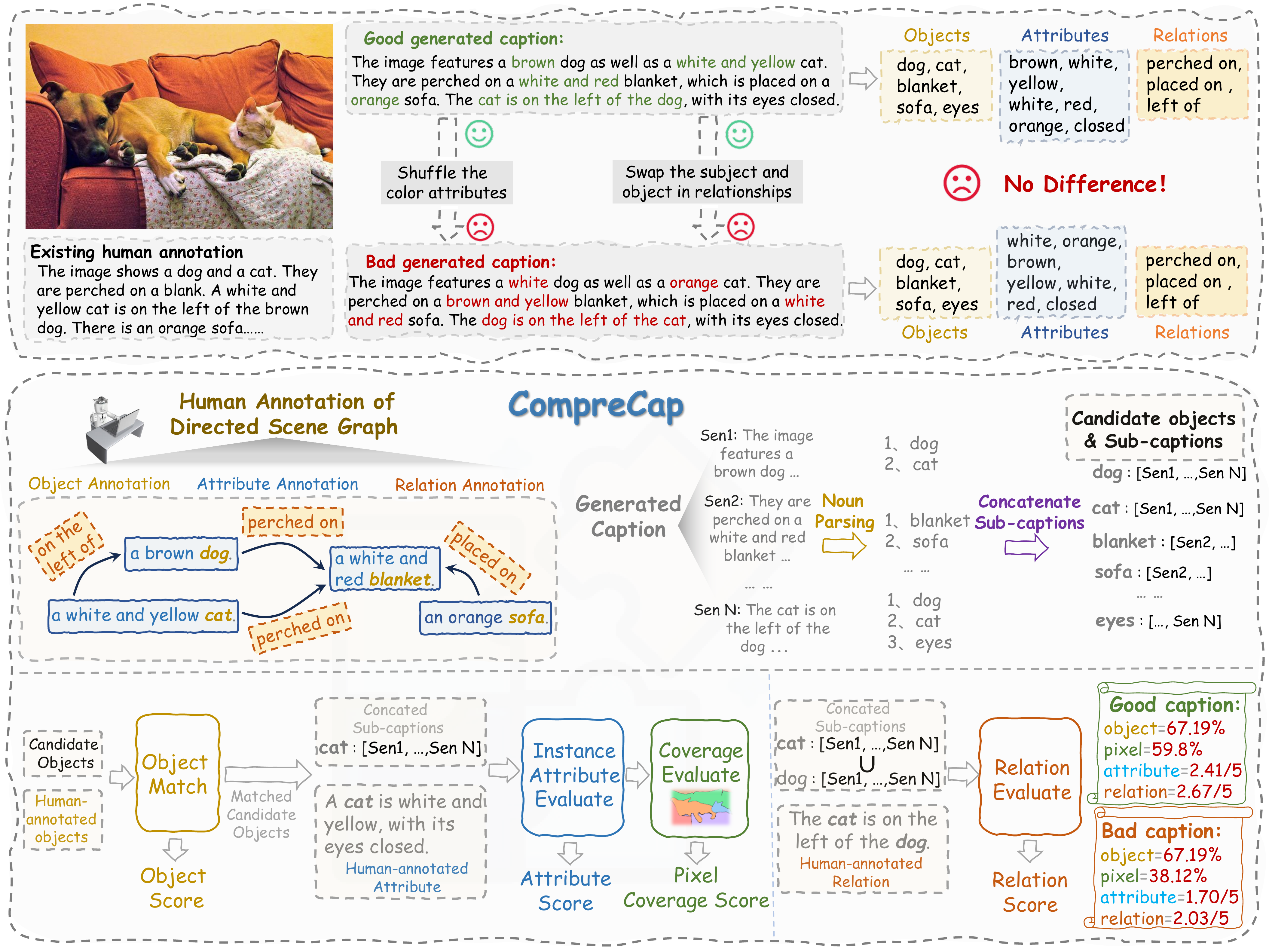}
    \vspace{-3mm}
    \caption{In the absence of the annotation of the directed scene graph, the extraction of isolated terms of objects, attributes, and relations disrupts the structured scene graph information, resulting in an inability to distinguish the quality of generated captions. We construct the \method which is manually annotated with the attributes bound to objects and the directional relationships between objects. Based on the high-quality \method dataset, we decompose the generated captions into a hierarchical structure and align them with object, attribute, and relation level annotations, enabling a more solid assessment of detailed captions.}
    \label{fig:motivation}
    \vspace{-5mm}
\end{figure*}

Image captioning is a fundamental task in the vision-language field, aimed at articulating the visual contents of images through textual descriptions.
Several image captioning benchmarks have been established, including MSCOCO~\cite{lin2014microsoft} and NoCaps~\cite{agrawal2019nocaps}, where images are manually annotated with brief captions, averaging about 10 words as shown in~\cref{tab:data_composition}.
In contrast to the short captions, detailed captions which encompass more comprehensive visual content within images, have gained significant attention in the fields of multi-modal generation and understanding~\cite{DALL-E3, liu2024playground, chen2023sharegpt4v, lotlip, zhang2024long,zhai2024background}. 
With the emergence of Large Vision-Language Models (LVLMs)~\cite{liu2024visual,chen2023internvl,yang2023dawn}, these models are capable of efficiently generating a large number of detailed image captions.
%

%
However, the evaluation of comprehensive image captions generated by LVLMs is hindered by the short-caption annotations like MSCOCO~\cite{lin2014microsoft} and NoCaps~\cite{agrawal2019nocaps}, as these cursory captions limit including the full range of visual information in images. 
To better assess the quality of generated detailed captions, some hallucination benchmarks (\eg, POPE~\cite{li2023evaluating} and FGHE~\cite{wang2024mitigating}) only increase the number of object labels in the question-answer format.
Furthermore, DetailCaps~\cite{detailcaption} includes the attributes and relations into the evaluation, adopting text scene graph parser Factual~\cite{li2023factual} to obtain isolated object-level, attribute-level and relation-level words. However, just illustrated as~\cref{fig:motivation}, the individual matching of these elements undermines the inherent binding of attributes to objects and the directional relationships, leading to the inaccurate assessment of image captions. For example, although color attributes are shuffled and assigned to incorrect objects, their inherent qualities remain unchanged, which does not diminish the score of the generated caption.
%

As illustrated in~\cref{tab:data_composition},  there is an absence of a human-annotated benchmark for comprehensive image captioning that integrates objects, their associated attributes, and the directional relationships among them.
Inspired by the study of directed scene graph~\cite{chang2021comprehensive}, which is organized by these elements, we proposed the \method benchmark as illustrated in~\cref{fig:data_struct}.
%
Our \method encompasses a wide array of objects within images and provides attribute annotations that are tied to corresponding objects, along with relationship annotations that indicate the subject-object connections.

To construct the \method, we regard MSCOCO~\cite{lin2014microsoft} panoptic segmentation dataset as our data source.
However, the category labels and segmentation maps of this dataset are not precise enough. 
Thus, we compose a common-object categories vocabulary based on several well-known datasets~\cite{zhang2024recognize, lin2014microsoft, kuznetsova2020open, shao2019objects365,zhai2022exploring}, and provide new semantic segmentation annotations for common objects in images within this vocabulary.
We retained images in which the segmented regions account for more than 95\% of the total image area.
Subsequently, we manually annotate the common objects with detailed descriptions of the associated attributes and salient relationships with other objects, composing a complete and directed scene graph structure. 
During evaluation, we parse the generated detailed captions based on object nouns, and first calculate the object-level coverage. 
Then, we leverage the powerful Llama3~\cite{llama3} to assess each object's associated attribute descriptions and relationships with others, through the precise match with the attribute-level and relation-level annotations.
Moreover, we design a vision question answering (VQA) task focused on tiny objects, which occupy less than 5\% of total image pixels, to investigate the correlation between LVLMs' perceptual performance on fine-grained objects and their qualities of the generated detailed captions.

As illustrated in~\cref{fig:motivation}, our evaluation method provides more accurate quality differences between captions. We benchmark 10 popular LVLMs on \method dataset and implement a human study, where the human performances in both generating comprehensive captions and fine-grained objects VQA achieve the best, underscoring the rationality of \method. Additionally, experimental results on the high-quality \method dataset further confirm that our assessment methodology achieves a strong consistency with human evaluation scores across all LVLMs. 
\section{Related work}\label{sec:related}

\subsection{Large Vision-Language Models}
Large language models (LLMs), such as GPT-3~\cite{GPT-3}, Flan-T5~\cite{T5}, LLaMA~\cite{touvron2023llama} and Vicuna~\cite{vicuna}, have significantly advanced. Such progress also facilitates the development of multi-modal language models by integrating visual content into LLMs. InstructBLIP~\cite{dai2024instructblip} and MiniGPT4-v2~\cite{zhu2023minigpt} connect the EVA-ViT~\cite{fang2023eva} and LLM with QFormer~\cite{QFormer}. LLaVA-1.5~\cite{liu2024visual} replaces the QFormer~\cite{QFormer} with a simple MLP and ShareGPT4V-13B~\cite{chen2023sharegpt4v} improves the quality of training data base on it. To increase the resolution of input images, LLaVA-Next~\cite{liu2024llava} splits an image into four parts and inputs them into CLIP-ViT~\cite{radford2021learning}, while miniGemini-HD~\cite{li2024mini} introduces CLIP-ConvNext~\cite{liu2022convnet} to process high-resolution images. Recently, InternVL-Chat-V1-5~\cite{chen2023internvl} tries to close the gap to commercial multimodal models and has approached the performance of GPT4V~\cite{yang2023dawn} and Gemini Pro~\cite{team2023gemini}, with the support of a powerful ViT with 6B parameters.

\subsection{Image Caption Evaluation}
Traditional image captioning metrics, such as CIDER~\cite{vedantam2015cider}, BLUE~\cite{papineni2002bleu}, ROUGE-L~\cite{lin2004rouge}, and METEOR~\cite{banerjee2005meteor}, evaluate the generated captions based on n-grams techniques~\cite{brown1992class,zhai2023exploring}. These metrics can evaluate the quality of short captions, but they are sensitive to the coverage of the reference captions available, failing to stably assess long captions. CLIPScore and RefCLIPScore~\cite{hessel2021clipscore} use a pre-trained CLIP~\cite{radford2021learning} model to evaluate captions based on the image-text similarity, which are also not capable of evaluating long captions, considering CLIP is limited to processing text within 77 tokens. 
DetailCaps~\cite{detailcaption} released a 100-case human-annotated detailed image captioning benchmark called DetailCaps-100, which contains an average of 177 words per image. It adopted text scene graph parser Factual~\cite{li2023factual} to extract object-level, attribute-level and relation-level texts from captions, and align these elements individually to get a score. However, parsing isolated scene graph components from long captions disrupts the binding nature of attributes to objects and the directional relations.
We construct the detailed image captioning benchmark \method, which incorporates human-annotated scene graphs that integrate objects, attributes bound to objects, and directional relations between objects. With the \method dataset, we comprehensively evaluate the quality of generated detailed captions based on object-level coverage, the accuracy of attribute descriptions, and the capture of key relations.

\section{CompreCap}\label{sec:method}

A comprehensive image caption should capture the existence of as many objects as possible, while also describing the attributes and relationships associated with the objects.
Based on this purpose, we propose a novel comprehensive image captioning benchmark called \method, with manually annotated scene graphs.
In~\cref{sec:scene-graph}, we present the directed scene graph annotation process of \method. Moreover, the construction of fine-grained object visual question answering (VQA) is presented in~\cref{sec:QA-anno}. We elaborated the comprehensive evaluation on generated detailed captions from LVLMs in~\cref{sec:long_caption_eval}.


\subsection{Scene Graph Annotation}
\label{sec:scene-graph}
The widely-used MSCOCO~\cite{lin2014microsoft} panoptic segmentation dataset is selected as the source dataset.
However, the segmentation maps and category labels of this dataset are not precise enough.
Moreover, it lacks descriptions of bound attributes and directional relations. 
Thus, we manually improve and extend the annotation of this dataset to construct a complete scene graph structure shown as~\cref{fig:data_struct}, for the evaluation of comprehensive image captions.

\noindent\textbf{Object Annotation.}
To cover as many objects as possible, we first collect a common categories vocabulary from several well-known datasets including RAM~\cite{zhang2024recognize}, COCO~\cite{lin2014microsoft}, OpenImagesV4~\cite{kuznetsova2020open}, and Object365~\cite{shao2019objects365}. 
Then, we re-annotate category labels and more precise segmentation maps for common objects in images within the vocabulary. 
To ensure completeness of the re-annotated objects within images, we exclude images where segmented areas cover less than 95\% of the total area.
%
%
Ultimately, \method comprises 560 images, with an average mask coverage of 95.83\% and a total of 412 object categories.

\noindent\textbf{Attribute Annotation.}
With the annotation of objects, we assign a detailed attribute annotation to each individual labeled object, to describe their salient visual characteristics including shape, size, color, and texture, \textit{etc}.  
To avoid confusion among different objects of the same category within an image during evaluation, annotators carefully labeled them with distinct attribute descriptions.

\noindent\textbf{Relation Annotation.} 
Besides the attribute annotations, we manually annotate the significant relationships between objects within the images.
The relation annotations encompass complete subject-verb-object structures, indicating the directional nature of relations between objects. 
In addition to spatial orientations such as above, below, left, and right, these relations also include verbs.

\subsection{Question-Answering Annotation}
\label{sec:QA-anno}
During the assessment of detailed captions, we observe LVLMs often ignore tiny objects, which occupy less than 5\% pixels of the global image. 
Hence, we design VQA tasks for these objects in a choice format, to analyze the correlation between LVLMs’ perceptual performance on fine-grained objects and the qualities of the generated detailed captions.
Specifically, we construct the CompreQA-for-presence (CompreQA-P) and the CompreQA-for-caption (CompreQA-Cap) tasks.
%

In CompreQA-P, we query LVLMs whether the presence of tiny objects can be discerned in an image. 
To overcome the potential inclination of LVLMs to habitually respond with `Yes, I can see it.' instead of actually perceiving this object, we additionally introduced an equivalent number of question-answer pairs for objects not present in the image as hallucination components. 
We evaluate LVLMs' ability to select the correct description for tiny objects in  CompreQA-Cap, which provides options for one correct caption and two incorrect captions. The attribute annotations of fine-grained objects are set as the answers, and we generated erroneous captions with the aid of GPT4~\cite{gpt4}. 
%
%
We manually modify the generated wrong captions to ensure that they conflict with the actual visual contents. 
The unrealistic ones (\eg, `a horse swimming under the water') are also revised. 
The correct options in both CompreQA-P and CompreQA-Cap have been randomly shuffled, and the examples of them are displayed in \supp.

\begin{figure}[t]
    \centering
    \captionsetup{skip=5pt} 
    \caption{We utilize the prompt to guide LLM to analyze whether the sub-captions contain a similar concept with the given phrase and provide a score from 0 to 5.}
    \label{fig:prompt}

    \begin{minipage}{1.0\columnwidth}
    \centering
    \begin{tcolorbox} 
        \centering
        \small
         \hspace{-6mm}
    \begin{minipage}{0.99\columnwidth}
    messages = [ \\
    \{\texttt{"role":"system", "content":} \\ f\texttt{"}You are a Natural Language Processing (NLP) expert. A curious human will give you a sentence and a phrase. The human wants you to help with analyzing whether the sentence includes a similar concept with the given phrase and rate the similarity on a scale from 0 to 5, with 0 being `completely lacks similar concepts' and 5 being `extremely has similar concepts'. You need to give helpful and reliable answers to help the human. \texttt{"}\},\\
    \{\texttt{"role":"user", "content":} \\ f\texttt{"}Sentence: $\langle$Concated sub-captions$\rangle$. Phrase: $\langle$Attribute annotations$\rangle$. Please provide an integer score as a single number from 0 to 5 without explanation.
    \texttt{"}\}
    \\]
    \end{minipage}
    \end{tcolorbox}
    \end{minipage}
    \vspace{-5pt} 
\end{figure}

\subsection{Structured Evaluation on Detailed Caption}
\label{sec:long_caption_eval}

%
Given the generated detailed captions from LVLMs, we will evaluate them from a directed scene graph view, which encompasses the objects along with their bound attributes and directional relations.
Moreover, with the re-annotated segmentation maps, we assess the pixel coverage of objects accurately mentioned in the captions.

\noindent\textbf{Decomposing Detailed Caption.} 
With the human-annotated directed scene graph which encompasses the descriptions at the levels of objects, attributes and relations, we parse the generated detailed captions into a hierarchical structure. This standard format allows for a more accurate match with the annotations and improves the stabilization of evaluation results.
Specifically, we first split the generated caption into several sub-captions by identifying sentence separators like periods. 
Then, we employ the noun extractor spaCy~\cite{honnibal2020industrial} in these sub-captions for noun extraction and lemmatization.
To this end, a set of candidate objects from generated detailed captions has been prepared.
%


\noindent\textbf{Object-level Evaluation.} 
In this part, we focus on the detecting accuracy of common semantic objects.
%
The matching scores between the extracted candidate objects and human-annotated objects are first calculated. 
%
%
Specifically, we use the text feature extractor Sentence BERT~\cite{reimers2019sentence} to obtain word embeddings and compute the similarity matrix $\mS \in \sR^{n\times m}$ between $n$ candidate objects and $m$ human annotated categories. 
%
Then, the candidate objects with the maximum similarity across both rows and columns in the $\mS$ are viewed to cover the corresponding human-annotated objects. The coverage of these candidate objects is recorded in $\mS^{'}$, where the elements are represented as:
\vspace{-0.6em}
\begin{equation}\label{e:objectcover} 
    \emS_{i,j}^{'}  = 
        \begin{cases} 
        \emS_{i,j}, \text{if } \mathop{\arg\max}\limits_{i^{'}} \ \mS_{i^{'},j}=i \text{ and } \mathop{\arg\max}\limits_{j^{'}} \ \mS_{i,j^{'}}=j, \\
        0, \text{otherwise}.
        \end{cases}
        \nonumber
\vspace{-0.35em} 
\end{equation}
The object-level coverage $S_\text{object}$ is defined as $\sum\limits_{i,j}\emS_{i,j}^{'}/ |\mS^{'}|$, denoting the soft coverage of the annotated categories of objects been mentioned in the generated captions.


\noindent\textbf{Attribute-level Evaluation.} 
Object-level coverage $S_{\text{object}}$ measures the capture of semantic objects in the generated detailed caption, but this metric does not focus on their descriptions of attributes.
Thus, we construct the attribute-level evaluation for the annotated objects mentioned by LVLMs, whose coverages in $\mS^{'}$ are not equal to 0. 
We find that an annotated object might be mentioned multiple times in the generated caption (\eg, boys as shown in~\cref{fig:longcaption_eval_sample}), so we split the generated detail captions with periods, and concatenate the sub-captions containing the same annotated object together.
Then, we adopt the well-designed prompt shown in~\cref{fig:prompt} to require a LLM to score the concatenated captions corresponding to the $i$-$th$ annotated object mentioned by LVLMs from 0 to 5 based on its human-annotated attribute phrases, and obtain the score matrix $\mA$, where $\emA_{i} \in \{0, 1, 2, 3, 4, 5\}$.
The attribute-level score is defined as $S_{\text{attribute}}=\sum\limits_{i}\emA_{i}/ |\mA|$ that is the average score of bound attribute descriptions of the annotated objects that have been covered in the generated captions.
Moreover, with the $\emA_{i}$ and segmentation maps, we can easily output the pixel-level coverage of attribute-related objects in the global image.
The soft pixel coverage of each image is computed as $\text{S-Cov}$$=$$(\sum\limits_{i}\emA_{i}\cdot \emM_{i})/5$, where $\emM_{i}$ is the percentage of pixels of the $i$-$th$ annotated object mentioned by LVLMs within the entire image.

%
%

\noindent\textbf{Relation-level Evaluation.}
The directional relation annotations involve two objects to represent subject-object correlation. Therefore, we connect the generated sub-captions that contain all objects in the relation annotations. Then, we implement the relation-level evaluation for objects following the same process of assessing attribute descriptions with an LLM. The relation-level score $S_{\text{relation}}$ measures the quality of relation descriptions of the annotated objects that have been covered in the generated captions.

\noindent\textbf{Unified Metric.}
To measure the overall quality of image captions, we employ the weighted average of $S_{\text{object}}$, $S_{\text{attribute}}$ and $S_{\text{relation}}$ as the unified metric:
\vspace{-0.6em}
\begin{equation}\label{e:unified} 
S_{\text{unified}} = \gamma {\widetilde{S}_{\text{object}}} + \lambda {\widetilde{S}_{\text{attribute}}} + \mu {\widetilde{S}_{\text{relation}}},
\nonumber
\vspace{-0.35em} 
\end{equation}
where $\widetilde{S}$ denotes that we linearly scale $S_{\text{object}}$, $S_{\text{attribute}}$ and $S_{\text{relation}}$ to the range of 0 to 100. Given the varying difficulty in describing objects, attributes, and relationships, we assign greater weight to the more challenging components: 25\%, 35\%, and 40\%, respectively.


\section{Experiments}\label{sec:exp}

\subsection{Experimental Setups}

\begin{table*}[t]
    \captionsetup{skip=5pt}
    \caption{Evaluation the detailed captions generated by the 10 LVLMs on \method benchmark. The \textbf{best} results are highlighted in bold. The \underline{second} and \underline{\underline{third}} best results are highlighted in underline and double underline, respectively. The human performance is also reported and it surpasses all LVLMs across every metric.}
    \label{tab:longcap_main}
    \centering
    \scriptsize
    \SetTblrInner{rowsep=1.pt}      
    \SetTblrInner{colsep=2.pt}      
    \resizebox{\linewidth}{!}{
    \begin{tblr}{
        cells={halign=c,valign=m},   
        hline{1,2,13}={1.0pt},         
        hline{10,12}={},
        vline{2,4,5}={},         
        row{12}={c,gray!20},
    }
        Model & Visual Encoder & LLM  &{ Caption\\ Length}&\( S_{\text{object}} (\%)\uparrow\) & { $S_{\text{attribute}}$ \(\uparrow\) \\ (\text{0}$\sim$\text{5}) } & { $S_{\text{relation}}$ \(\uparrow\) \\ (\text{0}$\sim$\text{5}) } & \( \text{S-Cov} (\%)\uparrow\)  & { $S_{\text{unified}}$ \(\uparrow\) \\ (\text{0}$\sim$\text{100}) } \\
        InstructBLIP-7B~\cite{dai2024instructblip} & \scriptsize{EVA-ViT-G} & Vicuna-7B & 69.93& 56.20& 1.89\std{0.00}& 2.53\std{0.00}& 42.03\std{0.14} & 48.16 \\
        MiniGPT4-v2~\cite{zhu2023minigpt} & EVA-ViT-G & Llama2-Chat-7B & 350.42& 56.74& 1.86\std{0.00}& 1.88\std{0.01}& 43.03\std{0.19} & 42.28 \\
        LLaVA-1.5-13B~\cite{liu2024visual} & CLIP-ViT-L/14& Vicuna-13B&86.97& 59.86& 2.01\std{0.01}& 2.59\std{0.00}& 43.81\std{0.25} & 50.32 \\
        ShareGPT4V-13B~\cite{chen2023sharegpt4v}& CLIP-ViT-L/14 & Vicuna-13B & 155.91& 67.88& 2.40\std{0.01}& 2.69\std{0.00}& 55.86\std{0.17} & 55.56 \\
        LLaVA-Next-llama3-8B~\cite{liu2024llava} &CLIP-ViT-L/14 & Llama3-8B-Instruct &  168.99& 70.22& 2.48\std{0.00}& 2.72\std{0.01}& 56.95\std{0.08} & 56.91 \\
        miniGemini-HD-34B~\cite{li2024mini}&{CLIP-ViT-L/14 \&\\CLIP-ConvNext-L} &Yi-34B & 173.71& 70.70& 2.48\std{0.00}& 2.70\std{0.00}& 57.20\std{0.11} & 56.88 \\
        InternVL-Chat-V1-5~\cite{chen2023internvl} &InternViT-6B-V1-5 &InternLM2-Chat-20B &115.22& 70.56& 2.50\std{0.00}& \underline{2.87}\std{0.00}& \underline{57.58}\std{0.09} & \underline{\underline{58.48}} \\
        LLaVA-Next-34B~\cite{liu2024llava}& CLIP-ViT-L/14 & Yi-34B &179.24& {\bf72.86}& {\bf2.59}\std{0.00}& \underline{\underline{2.79}}\std{0.00}& {\bf58.49}\std{0.15} & \underline{58.85} \\
        GPT-4V~\cite{yang2023dawn} &- &- &202.06& \underline{\underline{72.31}}& \underline{\underline{2.52}}\std{0.00}& 2.73\std{0.00}& 57.27\std{0.14} & 57.74 \\
        
        GPT-4o~\cite{gpt4o} &- &- & 108.20& \underline{72.78}& \underline{2.58}\std{0.00}& {\bf2.93}\std{0.00}& \underline{\underline{57.54}}\std{0.23}  & {\bf60.05} \\

        Human &- &- & 133.61& {\color{red}\textbf{77.62}}& {\color{red}\textbf{2.78}}\std{0.00}& {\color{red}\textbf{2.99}}\std{0.00}& {\color{red}\textbf{59.58}}\std{0.16} & {\color{red}\textbf{62.99}} \\

        \end{tblr}
        }
\vspace{-1.5mm}
\end{table*}

\begin{table}[!ht]
    \captionsetup{skip=5pt}
    \caption{Our method achieves a strong consistency with human judgment. Our annotations of the directed scene graph facilitate more accurate
    matching at the levels of objects, attributes and relations during
    comprehensive caption assessment. $S_{\text{Llama3}}$ denotes the unified metric directly produced by Llama3.}
    \centering\small
    \setlength{\tabcolsep}{5.0pt}
    \begin{tabular}{c|c|c|c}
    \hline
     & Person r $\uparrow$ & Kendall’s $\tau$ $\uparrow$ & Spearman's $\rho$ $\uparrow$ \\\hline
    BLEU@4~\cite{papineni2002bleu} & 0.867 & 0.563 & 0.763  \\
    METEOR~\cite{banerjee2005meteor} & 0.581 & 0.330 & 0.423 \\
    ROUGE-L~\cite{lin2004rouge} & 0.725 & 0.331  & 0.428 \\
    CIDER~\cite{vedantam2015cider} & 0.563 & 0.403 & 0.542  \\
    CLIPScore~\cite{hessel2021clipscore} & 0.691 & 0.426 & 0.539 \\\hline
    $S_{\text{Llama3}}$ & 0.816 & 0.309  & 0.445 \\\hline
    \rowcolor{lightgray!44} $S_{\text{unified}}$ & \textbf{0.989} & \textbf{0.927} & \textbf{0.982}  \\
    \hline
    \end{tabular}
\vspace{-12pt} 
    \label{tab:quan_consis}
\end{table}

\noindent\textbf{LVLMs for Evaluation.}
Based on \method, we evaluate 10 Large Vision-Language Models (LVLMs) including 8 classic and popular open-source models and 2 close-source models, \ie, InstructBLIP-7B~\cite{dai2024instructblip}, MiniGPT4-v2~\cite{zhu2023minigpt}, LLaVA-1.5-13B~\cite{liu2024visual}, ShareGPT4V-13B~\cite{chen2023sharegpt4v}, LLaVA-Next-llama3-8B~\cite{liu2024llava}, InternVL-Chat-V1-5~\cite{chen2023internvl}, miniGemini-HD-34B~\cite{li2024mini}, LLaVA-Next-34B~\cite{liu2024llava}, GPT-4V~\cite{yang2023dawn} and GPT-4o~\cite{gpt4o}. We evaluate the detailed captions generated by these VLMs at the object level, attribute level, and relation level. Moreover, we evaluate the fine-grained object perception of these VLMs with the designed visual question-answering task.

\begin{figure*}[t]
    \centering
    \includegraphics[width=0.99\textwidth]{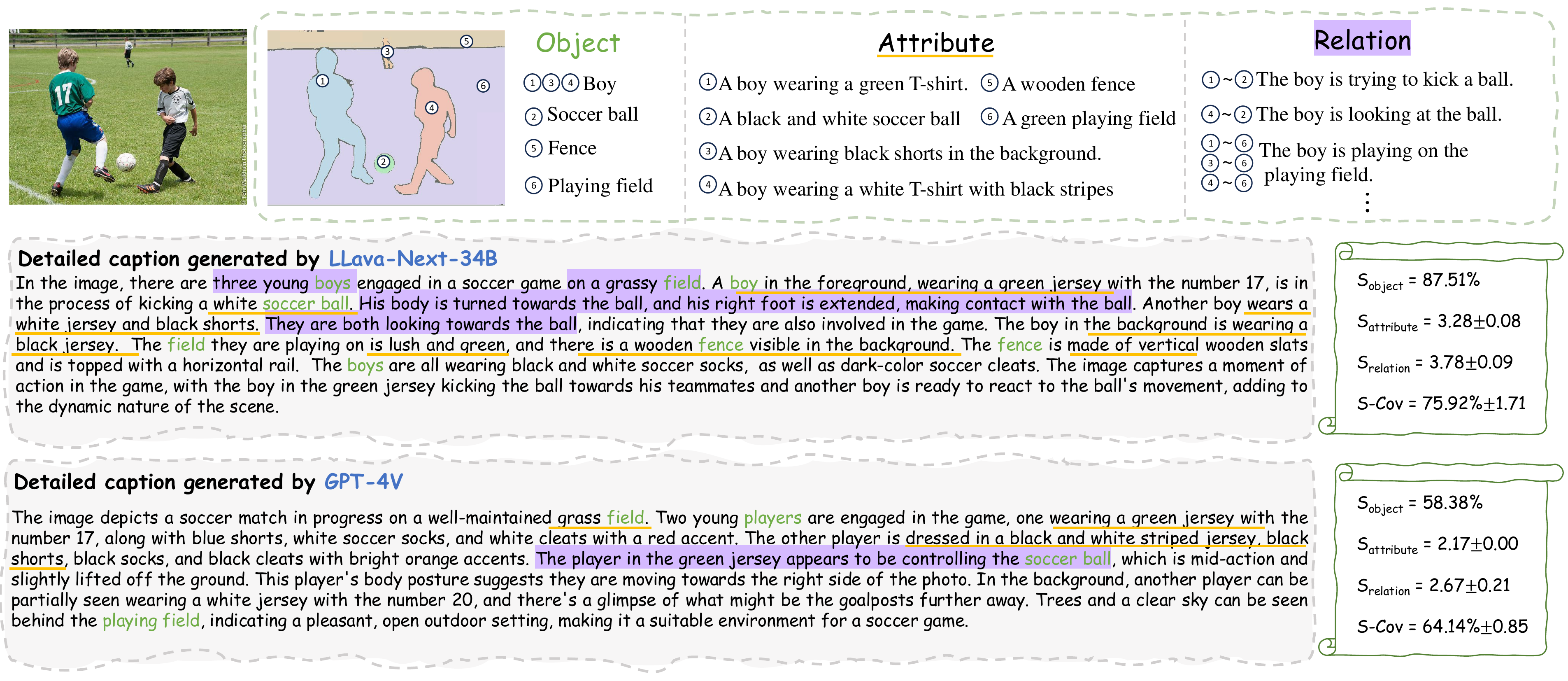}
    \vspace{-4mm}
    \caption{Illustration of a comprehensive caption evaluation example. The words or sub-captions corresponding to the human annotated object, attribute, and relation are in green, underlined in yellow, and highlighted in purple, respectively.
    GPT-4V fails to identify objects such as the `fence' in the background compared to LLaVA-Next-34B~\cite{liu2024llava}. Additionally, the attribute and relationship description from LLaVA-Next-34B are more consistent with human annotations.}
    \label{fig:longcaption_eval_sample}
\vspace{-4mm} 
\end{figure*}

\begin{figure}[htbp]
\centering
    \subfloat[$S_{\text{unified}}$ directly output by Llama3]{
         \centering
         \includegraphics[width=0.88\linewidth]{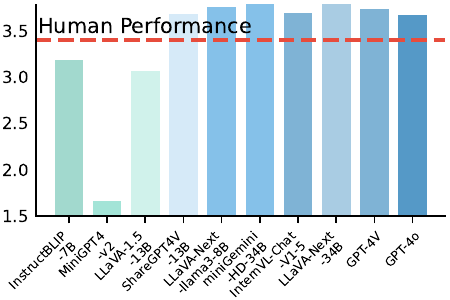}
         \vspace{-1mm}
    }\\
    \subfloat[$S_{\text{unified}}$ from \textbf{our method}]{
         \centering
         \includegraphics[width=0.88\linewidth]{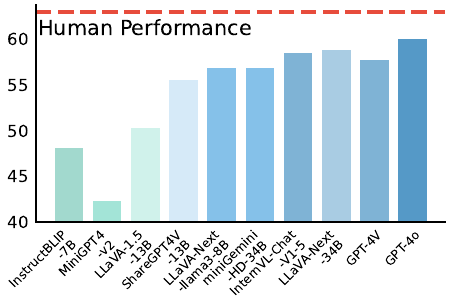}
         \vspace{-1mm}
    }
    \captionsetup{skip=3pt}
    \caption{The human performance exceeds all LVLMs with our evaluation method. Our annotations provide more precise references when assessing detailed captions.}
\vspace{-5mm}
\label{fig:ablate_nvs}
\end{figure}

\noindent\textbf{Implementation details.}
For evaluating comprehensive captions, we first guide LVLMs to generate detailed captions of the images in \method dataset, with their default hyper-parameters and the prompt `Please describe the image in detail, focusing on the visible objects and the relationships among these objects.'.
Then we evaluate the quality of these generated captions following the steps present in~\cref{sec:long_caption_eval}. LLama3~\cite{llama3} is employed as an LLM evaluator, considering it is known as the best existing open-source LLM. 
We utilize NVIDIA A100 GPU for the generation and evaluation of detailed captions. Since there are slight fluctuations in the predictions of LLM and LVLMs, we repeat the evaluation three times and report the mean and standard deviation of the results.

\begin{figure*}[t]
\centering 
\captionsetup{skip=3pt}
\subfloat[LLaVA-Next-34B]{
     \centering
     \includegraphics[width=0.31\linewidth]{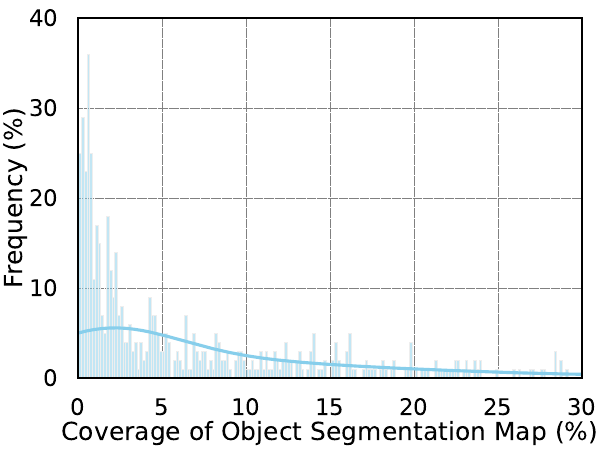}
 }
 \hfill
\subfloat[InternVL-Chat-V1-5]{
     \centering
     \includegraphics[width=0.31\linewidth]{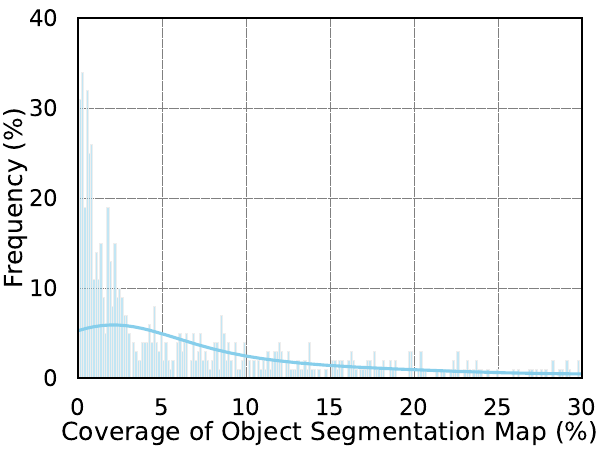}
}
 \hfill
\subfloat[GPT4V]{
     \centering
     \includegraphics[width=0.31\linewidth]{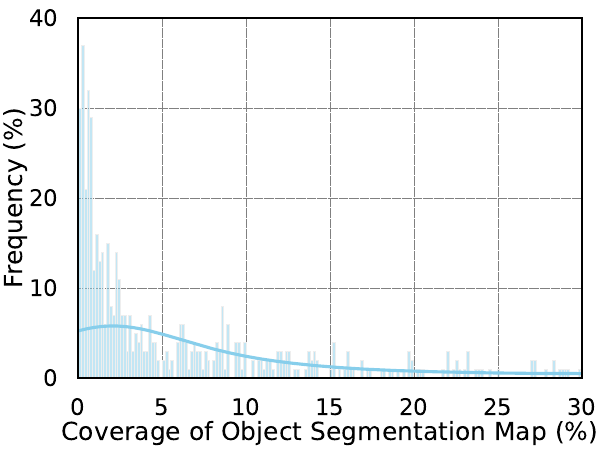}
}
\caption{We calculate the pixel area of the ignored objects in generated captions, and present the top 3 LVLMs on the performance of the S-Cov metric. These LVLMs tend to ignore fine-grained objects (segmentation map coverage \textless 5\%) when describing the given image.}
\label{fig:region_zero}
\vspace{-3mm}
\end{figure*}

\subsection{Comprehensive Image Caption Evaluation}
\noindent\textbf{Analysis of the Results.}
We evaluate the comprehensive captions generated by 10 LVLMs and report the results in~\cref{tab:longcap_main}. The average lengths of the generated captions from different models vary from 69.93 to 359.42.
In general, LLaVA-Next-34B~\cite{liu2024llava} and GPT-4o~\cite{gpt4o} achieve the best performance, with the former slightly outperforming the latter in terms of object coverage, attribute scores, and pixel coverage, but it lags behind by 0.14 in relation scores on a five-point scale. However, GPT-4o~\cite{gpt4o} reaches comparable results using more concise text, generating about 60\% of the output length of LLaVA-Next-34B~\cite{liu2024llava}. Similarly, we find that the quality of captions at the levels of object, attributes and relationship does not correlate with caption length. For example, while captions generated by MiniGPT4-v2 are nearly twice the average length of those produced by LLaVA-Next-34B~\cite{liu2024llava}, their corresponding quality metrics are significantly lower than those of LLaVA-Next-34B~\cite{liu2024llava}.
As the well-known best model before GPT-4o~\cite{gpt4o}, GPT-4V~\cite{yang2023dawn} falls short against LLaVA-Next-34B~\cite{liu2024llava} in generating comprehensive image captions. To analyze the gap between them, we demonstrate an evaluation example in~\cref{fig:longcaption_eval_sample}. The instance illustrates that LLaVA-Next-34B~\cite{liu2024llava} captures more objects and provides a more detailed description of attributes and relationships.

Moreover, we assess the human performance based on the human descriptions of the images in \method. The superior performance of humans compared to all LVLMs shown in~\cref{tab:longcap_main} validates the integrity of annotations and the rationale evaluation method in \method.

\noindent\textbf{Comparison with traditional metrics.}
To demonstrate the credibility of our \method benchmark,
we quantitatively compare the human judgment consistency of our $S_{\text{unified}}$ with traditional captioning metrics across model and human performances in~\cref{tab:quan_consis}. The comparison metrics encompass n-gram-based methods~\cite{papineni2002bleu,banerjee2005meteor,lin2004rouge,vedantam2015cider} and CLIPScore~\cite{hessel2021clipscore}. Due to the limitations of n-gram algorithms in modeling long contextual dependencies or the constraint of a maximum of 77 tokens, these metrics are unable to effectively capture and evaluate the various elements in detailed captions. In contrast, our approach maintains strong consistency with human scoring, demonstrating the reliability and practicality of our \method in evaluating detailed image captions.
The human evaluation scores and more comparisons about human consistency are included in \supp.

\noindent\textbf{Effect of the Directed Scene Graph Annotation.}
With the annotated directed scene graph of \method, we can decouple the generated captions into a hierarchical structure, allowing for a more precise match with annotations at the levels of objects, attributes, and relationships. To validate the efficacy of our scene graph annotations, we compare our method with directly scoring the generated detailed captions by Llama3~\cite{llama3}, focusing on human performance and consistency with human evaluation scores. We first organize the annotations of objects, attributes and relations in \method into coherent ground-truth (GT) captions. Then we require Llama3~\cite{llama3} to output a unified score ranging from 0 to 5 by comprehensively considering the quality of objects, attributes, and relations based on the GT caption.
The complete results of the direct evaluation with Llmam3 are shown in \supp.

As revealed in~\cref{fig:ablate_nvs}, when the high-quality captions generated by human are directly scored using Llama3~\cite{llama3}, their scores are lower than those of most LVLMs, which is unexpected. In contrast, human performance from our approach achieves significant superiority, validating the comprehensive annotations of \method and the rationality of our evaluation method.
%
Moreover, our $S_{\text{unified}}$ faithfully aligns with human evaluation
scores compared to the result directly obtained from Llama3, shown as the last two rows of~\cref{tab:quan_consis}. The high consistency with human judgment emphasizes the credibility and practicality of our
\method, proving its promising prospects in LVLMs evaluation.  The comparisons on scores of objects, attributes and relations can be seen in \supp.

\subsection{Fine-grained Object VQA}
We show the proportion of pixels of objects omitted in the generated captions in~\cref{fig:region_zero}. Most of these ignored objects are tiny, the segmentation map of which only occupies less than 5\% of the global image. 
To analyze the correlation between LVLMs’ perception of fine-grained objects and their qualities of generating detailed captions, we evaluate the 9 LVLMs on the fine-grained object VQA, including CompreQA-P and CompreQA-Cap. The evaluation for GPT-4V is lacking because its API has been deprecated after annotating the final versions of this VQA. The results in \cref{tab:VQA_result} show that InternVL-Chat-V1-5~\cite{chen2023internvl} achieves the best performance among LVLMs benefiting from a powerful visual encoder with 6B parameters. LLaVA-Next-34B~\cite{liu2024llava} and GPT-4o~\cite{gpt4o} rank second and third, respectively. In CompreQA-P, the hallucination created by non-existent objects results in InstructBLIP-7B~\cite{dai2024instructblip} and MiniGPT4-v2~\cite{zhu2023minigpt} only achieving the accuracy rates comparable to, or even worse than random guessing (\ie, 50\%). The human performance also surpasses all LVLMs on both CompreQA-P and CompreQA-Cap, with an accuracy over 96\%, underscoring the rationality of the fine-grained object VQA task in \method. In general, the ranking of LVLMs on this VQA task aligns with their ranking in terms of comprehensive caption quality.

\begin{table}[t]
\centering
\captionsetup{skip=5pt}
    \caption{Accuracy of LVLMs on CompreQA-P and CompreQA-Cap. The \textbf{best} 
             results are in bold. The \underline{second} and \underline{\underline{third}} best results are in underline and double underline, respectively. Human performance is also reported and surpasses all LVLMs.
        }
    \SetTblrInner{rowsep=1.pt}       
    \SetTblrInner{colsep=2.5pt}       
    \begin{tblr}{
        cells={halign=c,valign=m},   
        hline{1,2,12}={1.0pt},         
        hline{10,11}={},   
        vline{2}={},         
        row{11}={c,gray!20},
    }
            Model &  {\small{CompreQA-P} \\ \small{ACC}(\%) \(\uparrow\)} & {\small{CompreQA-Cap} \\\small{ACC}(\%) \(\uparrow\)}  \\
        InstructBLIP-7B~\cite{dai2024instructblip} & 35.28\std{0.00}& 36.52\std{0.00} \\
        MiniGPT4-v2~\cite{zhu2023minigpt} & 51.06\std{0.14}& 40.78\std{1.16} \\
        LLaVA-1.5-13B~\cite{liu2024visual} & 82.45\std{0.38}& 84.87\std{0.17} \\
        ShareGPT4V-13B~\cite{chen2023sharegpt4v}  & 82.03\std{0.44}& 85.34\std{0.33} \\
        miniGemini-34B-HD~\cite{li2024mini}  & 86.88\std{0.90}& 89.01\std{0.29} \\
        LLaVA-Next-llama3-8B~\cite{liu2024llava} & 88.48\std{0.52}& 90.90\std{0.17} \\
        InternVL-Chat-V1-5~\cite{chen2023internvl} & {\bf91.67}\std{0.00}& {\bf94.33}\std{0.00} \\
        LLaVA-Next-34B~\cite{liu2024llava} & \underline{91.43}\std{0.22}& \underline{92.55}\std{0.00} \\
        GPT-4o~\cite{gpt4o}  & \underline{\underline{90.96}}\std{0.38}& \underline{\underline{91.37}}\std{0.33} \\ 
        Human & {\color{red}\textbf{96.49}} &{\color{red}\textbf{96.83}} 

        \end{tblr}
\label{tab:VQA_result}
\vspace{-3mm}
\end{table}
\section{Conclusion}\label{sec:conclusion}
In this work, we construct the human-annotated \method benchmark with the directed scene graph, aiming to evaluate the comprehensive image captioning performance of LVLMs. 
The \method integrates a wide array of objects within images, the attribute descriptions bound to them, along with the directional relationships between corresponding objects.
Based on the comprehensive scene graph annotations, we develop an evaluation method to assess the detailed captions generated by 10 LVLMs through the precise match with the annotations at the levels of object, attribute and relationship. 
Experimental results show that the metrics obtained from our method faithfully align with human evaluation scores across various LVLMs, indicating the practicality and credibility of our \method.
The establishment of the \method benchmark contributes to a more in-depth assessment of LVLMs.

\section*{Acknowledgments}
This work is supported by the National Natural Science Foundation of China (NSFC) under Grants 62225207, 62436008, 62306295, 62476260 and the Fundamental Research Funds for the Central Universities under Grant WK2100000057.

\clearpage
\small
\bibliographystyle{ieeenat_fullname}
\bibliography{main}
\newpage
\appendix

\newcommand\DoToC{%
    \hypersetup{linkcolor=blue}
    \startcontents
    \printcontents{}{1}{\hrulefill\vskip0pt}
    \vskip0pt \noindent\hrulefill
    \hypersetup{linkcolor=red}
    }
\renewcommand\thesection{\Alph{section}}
\renewcommand\thefigure{S\arabic{figure}}
\renewcommand\thetable{S\arabic{table}}
\renewcommand\theequation{S\arabic{equation}}

\setcounter{figure}{0}
\setcounter{table}{0}
\setcounter{equation}{0}
\maketitlesupplementary
\section*{Appendix}

\noindent\DoToC 


\begin{table*}
    \caption{Analysis on the influence of image resolution and base LLM on the generated detailed captions. `Max Res.' denotes the maximum resolution.}
    \label{supp_tab:mllm_component}
    \centering
    \scriptsize
    \SetTblrInner{rowsep=1.pt}      
    \SetTblrInner{colsep=6.pt}      
    \resizebox{\linewidth}{!}{
    \begin{tblr}{
        cells={halign=c,valign=m},   
        hline{1,2,5}={1.0pt},         
        vline{2,4,5}={},         
    }
        Model & LLM & {Max Res.}  &{ Caption\\ Length}&\( S_{\text{object}} (\%)\uparrow\) &\( S_{\text{attribute}} \uparrow \) &\( S_{\text{relation}} \uparrow\)& \( \text{S-Cov.} (\%)\uparrow\)  \\

        LLaVA-1.5-13B~\cite{liu2024visual} & Vicuna-13B& 336& 86.97& 59.93& 2.02\std{0.01}& 2.60\std{0.00}& 44.11\std{0.01}\\
        LLava-Next-13B~\cite{liu2024llava} & Vicuna-13B& 1008 & 172.19& 70.55& 2.50\std{0.01}& 2.73\std{0.01}& 56.68\std{0.28}\\
        LLava-Next-34B~\cite{liu2024llava}& Yi-34B & 1008& 179.24& 72.86& 2.59\std{0.00}& 2.79\std{0.00}& {58.49}\std{0.15}  \\

        \end{tblr}
        }
\end{table*}

\begin{table*}[t]
\centering
    \caption{Accuracy of LVLMs on CompreQA-P and CompreQA-Cap. The \textbf{best} 
             results are in bold. The \underline{second} and \underline{\underline{third}} best results are in underline and double underline, respectively. `Max Res.' denotes the maximum resolution.
        }
    \SetTblrInner{rowsep=1.pt}       
    \SetTblrInner{colsep=2.5pt}       
    \begin{tblr}{
        cells={halign=c,valign=m},   
        hline{1,2,12}={1.0pt},         
        hline{10,11}={},   
        vline{2,5}={},         
    }
            Model & Visual Encoder & LLM & {Max Res.} &  {CompreQA-P \\ ACC(\%) \(\uparrow\)} & {CompreQA-Cap \\ ACC(\%) \(\uparrow\)} \\
        InstructBLIP-7B~\cite{dai2024instructblip} & EVA-ViT-G & Vicuna-7B & 224 & 35.28\std{0.00}& 36.52\std{0.00} \\
        MiniGPT4-v2~\cite{zhu2023minigpt}  & EVA-ViT-G & Llama2-Chat-7B  & 448 & 51.06\std{0.14}& 40.78\std{1.16}\\
        LLaVA-1.5-13B~\cite{liu2024visual}  & CLIP-ViT-L/14& Vicuna-13B & 336 & 82.45\std{0.38}& 84.87\std{0.17}\\
        LLava-Next-13B~\cite{liu2024llava} & CLIP-ViT-L/14& Vicuna-13B & 1008 & 84.81\std{0.59}& 81.68\std{0.44} \\
        ShareGPT4V-13B~\cite{chen2023sharegpt4v} & CLIP-ViT-L/14& Vicuna-13B & 336  & 82.03\std{0.44}& 85.34\std{0.33}
        \\
        miniGemini-34B-HD~\cite{li2024mini} &{CLIP-ViT-L/14 \&\\CLIP-ConvNext-L} &Yi-34B & 1536 & 86.88\std{0.90}& 89.01\std{0.29} \\
        LLaVA-Next-llama3-8B~\cite{liu2024llava} &CLIP-ViT-L/14 & Llama3-8B-Instruct & 1008 & 88.48\std{0.52}& 90.90\std{0.17}\\
        InternVL-Chat-V1-5~\cite{chen2023internvl} &InternViT-6B-V1-5 &InternLM2-Chat-20B & 448 & {\bf91.67}\std{0.00}& {\bf94.33}\std{0.00} \\
        LLaVA-Next-34B~\cite{liu2024llava}& CLIP-ViT-L/14 & Yi-34B &  1008 & \underline{91.43}\std{0.22}& \underline{92.55}\std{0.00}\\
        GPT-4o~\cite{gpt4o} &- &-  &- & \underline{\underline{90.96}}\std{0.38}& \underline{\underline{91.37}}\std{0.33}
        \end{tblr}
\label{supp_tab:mllm_component_vqa}
\end{table*}

\section{More Experimental Analysis}
\label{supp_sec:additional_analysis}

\begin{table*}[t]
    \caption{Evaluation of the detailed captions generated by the 10 LVLMs on \method benchmark, including the human evaluation score reported. The \textbf{best} results are highlighted in bold. The \underline{second} and \underline{\underline{third}} best results are highlighted in underline and double underline, respectively.}
    \label{supp_tab:longcap_main_whumanscoring}
    \centering
    \scriptsize
    \SetTblrInner{rowsep=1.5pt}      
    \SetTblrInner{colsep=2.pt}      
    \resizebox{\linewidth}{!}{
    \begin{tblr}{
        cells={halign=c,valign=m},   
        hline{1,2,13}={1.0pt},         
        hline{10,12}={},
        vline{2,3,8}={},         
        row{12}={c,gray!20},
    }
        Model &{ Caption\\ Length}&\( S_{\text{object}} (\%)\uparrow\) &\( S_{\text{attribute}} \uparrow \) &\( S_{\text{relation}} \uparrow\)& \( \text{S-Cov.} (\%)\uparrow\) & \( S_{\text{unified}}\uparrow \)  & {Human \\ Evaluation} \\

        InstructBLIP-7B~\cite{dai2024instructblip} & 69.93& 56.20& 1.89\std{0.00}& 2.53\std{0.00}& 42.03\std{0.14} & 48.16 & 2.36\\
        MiniGPT4-v2~\cite{zhu2023minigpt} & 350.42& 56.74& 1.86\std{0.00}& 1.88\std{0.01}& 43.03\std{0.19} & 42.28 & 1.83 \\
        LLaVA-1.5-13B~\cite{liu2024visual} & 86.97& 59.86& 2.01\std{0.01}& 2.59\std{0.00}& 43.81\std{0.25} & 50.32 &  2.39 \\
        ShareGPT4V-13B~\cite{chen2023sharegpt4v} & 155.91& 67.88& 2.40\std{0.01}& 2.69\std{0.00}& 55.86\std{0.17} & 55.56 & 3.28 \\
        LLaVA-Next-llama3-8B~\cite{liu2024llava} &  168.99& 70.22& 2.48\std{0.00}& 2.72\std{0.01}& 56.95\std{0.08} & 56.91 & 3.34 \\
        miniGemini-HD-34B~\cite{li2024mini} & 173.71& 70.70& 2.48\std{0.00}& 2.70\std{0.00}& 57.20\std{0.11} & 56.88 & 3.37 \\
        InternVL-Chat-V1-5~\cite{chen2023internvl} &115.22& 70.56& 2.50\std{0.00}& \underline{2.87}\std{0.00}& \underline{57.58}\std{0.09} & \underline{\underline{58.48}} & 3.42 \\
        LLaVA-Next-34B~\cite{liu2024llava} &179.24& {\bf72.86}& {\bf2.59}\std{0.00}& \underline{\underline{2.79}}\std{0.00}& {\bf58.49}\std{0.15} & \underline{58.85} & 3.64 \\
        GPT-4V~\cite{yang2023dawn} &202.06& \underline{\underline{72.31}}& \underline{\underline{2.52}}\std{0.00}& 2.73\std{0.00}& 57.27\std{0.14} & 57.74 & 3.63 \\
        
        GPT-4o~\cite{gpt4o} & 108.20& \underline{72.78}& \underline{2.58}\std{0.00}& {\bf2.93}\std{0.00}& \underline{\underline{57.54}}\std{0.23} & {\bf60.05} & 3.68 \\

        Human & 133.61& {\color{red}\textbf{77.62}}& {\color{red}\textbf{2.78}}\std{0.00}& {\color{red}\textbf{2.99}}\std{0.00}& {\color{red}\textbf{59.58}}\std{0.16} & {\color{red}\textbf{62.99}} & 4.0 \\

        \end{tblr}
        }
\end{table*}

\subsection{Analysis on the effect of components of LVLMs }
\label{supp_sec:lvlm_componet}

We mainly analyze the effect of image resolution and base Large Language Model (LLM) on the quality of generated detailed captions and the perception of tiny objects in~\cref{supp_tab:mllm_component} and~\cref{supp_tab:mllm_component_vqa}, respectively.
Compared with LLaVA-1.5-13B~\cite{liu2024visual}, LLava-Next-13B~\cite{liu2024llava} increases the input image resolution.
The performance comparison of generating detailed captions between these two LVLMs indicates that increasing the input image resolution can enable the LVLM to recognize more objects, and describe the objects along with the key relations between them in a more precise way.
When upgrading the base LLM to the larger and more powerful Yi-34B~\cite{young2024yi}, LLava-Next-34B~\cite{liu2024llava} further widened its advantage over LLava-Next-13B~\cite{liu2024visual} in generating comprehensive captions as shown in~\cref{supp_tab:mllm_component}.
Moreover, we present the performance of 10 LVLMs in CompreQA-P and CompreQA-Cap shown as~\cref{supp_tab:mllm_component_vqa}. We compare LLaVA-1.5-13B~\cite{liu2024visual}, LLava-Next-13B~\cite{liu2024visual} and LLava-Next-34B~\cite{liu2024llava} and observe that gains brought by the base LLM have given LLaVA-Next-34B\cite{liu2024llava} a significant advantage on the two fine-grained objects VQA metrics. 
While increasing image resolution often improves fine-grained object understanding, we find that it is not the sole contributor to this task. For example, although the input resolution of InternVL-Chat-V1-5~\cite{chen2023internvl} is merely 448, not even one-third that of miniGemini-34B-HD~\cite{li2024mini}, the former surpasses the latter by 4.79\% and 3.54\% in two metrics, respectively.
The comparison between LLava-Next-13B~\cite{liu2024llava} and LLaVA-1.5-13B~\cite{liu2024visual} shows that the former features a several-fold increase in input resolution compared to the latter, and outperforms the latter on the CompreQA-P. However, LLava-Next-13B~\cite{liu2024llava} falls short on the CompreQA-Cap evaluation. This reaffirms that merely increasing input resolution is insufficient to enhance the model's comprehension of fine-grained objects.

\subsection{Consistency with human evaluation scores}
We present the complete results of human evaluation scores across 10 LVLMs and human performance in~\cref{supp_tab:longcap_main_whumanscoring}. All our metrics on \method dataset achieve a strong consistency with
human evaluation scores across all LVLMs and human performance, emphasizing the credibility of our \method and its promising prospects in LVLMs evaluation.

\subsection{Compare with directly scoring with Llama3}\label{supp_sec:compare_direct_llama}
\begin{figure*}[htbp]
\centering
    \subfloat[$S_{\text{object}}$ directly output by Llama3]{
         \centering
         \includegraphics[width=0.3\linewidth]{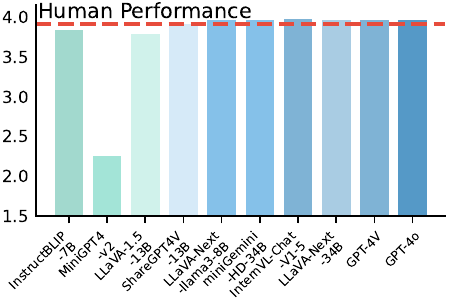}
         \vspace{-1mm}
    }
    \hfill
    \subfloat[$S_{\text{attribute}}$ directly output by Llama3]{
         \centering
         \includegraphics[width=0.3\linewidth]{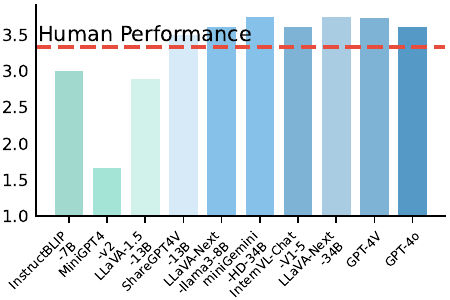}
         \vspace{-1mm}
    }
    \hfill
    \subfloat[$S_{\text{relation}}$ directly output by Llama3]{
         \centering
         \includegraphics[width=0.3\linewidth]{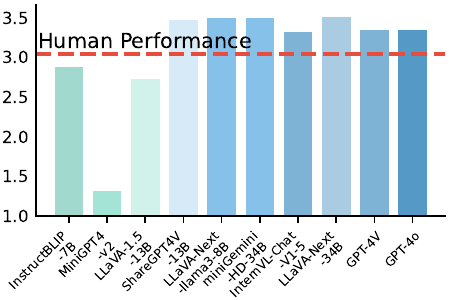}
         \vspace{-1mm}
    }\\
    \subfloat[$S_{\text{object}}$ from \textbf{our method}]{
         \centering
         \includegraphics[width=0.3\linewidth]{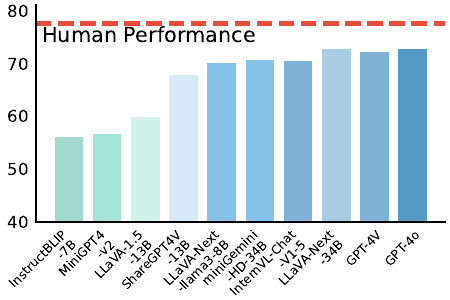}
         \vspace{-1mm}
    }
    \hfill
    \subfloat[$S_{\text{attribute}}$ from \textbf{our method}]{
         \centering
         \includegraphics[width=0.3\linewidth]{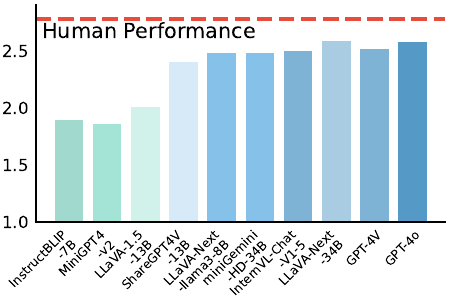}
         \vspace{-1mm}
    }
    \hfill
    \subfloat[$S_{\text{relation}}$ from \textbf{our method}]{
         \centering
         \includegraphics[width=0.3\linewidth]{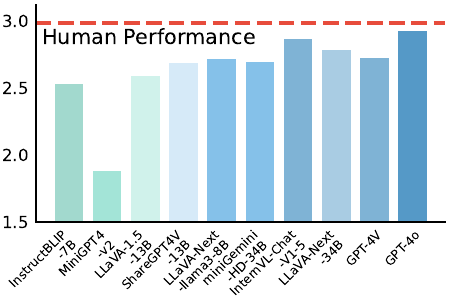}
         \vspace{-1mm}
    }
    \captionsetup{skip=5pt}
    \caption{The human performance exceed all LVLMs on the scores of 
             objects, attributes and relations with our evaluation method. Compared to directly scoring detailed captions with Llama3, the annotations of the directed scene graph provide more precise references across various dimensions.}
\label{supp_fig:ablate_nvs}
\end{figure*}

We first organize the annotated objects and the descriptions of attributes and relationships into coherent ground-truth (GT) captions. Then, we leverage Llama3~\cite{llama3} to directly score the generated detailed captions, with the prompt `Please quantify the quality of the given $\langle$\text{generated caption}$\rangle$ on a score scale from 0 to 5 for `object coverage', `object attributes', `relationships between objects' and `overall quality', without any other explanation, using the given $\langle$\text{GT caption}$\rangle$ as a standard. The higher the score, the more the given $\langle$\text{generated caption}$\rangle$ matches the content of the given $\langle$\text{GT caption}$\rangle$.'. Then we will obtain the object score, attribute score, relation score as well as an overall score. The results are shown in~\cref{supp_tab:llama_longcap_main_whumanscoring}. When directly assessing with Llama3~\cite{llama3}, the high-quality captions generated by human lag behind most LVLMs on scores of all dimensions, which is unexpected and unreasonable as shown in~\cref{supp_fig:ablate_nvs}. This indicates that with the annotated directed scene graph of \method, we can decouple the generated captions into a hierarchical structure, allowing for a more precise match with annotations at the levels of objects, attributes, and relationships.

\begin{figure}[htbp]
\centering
\captionsetup{skip=6pt}
    \subfloat[Directly scoring with Llama3]{
     \centering
     \includegraphics[width=0.49\linewidth]{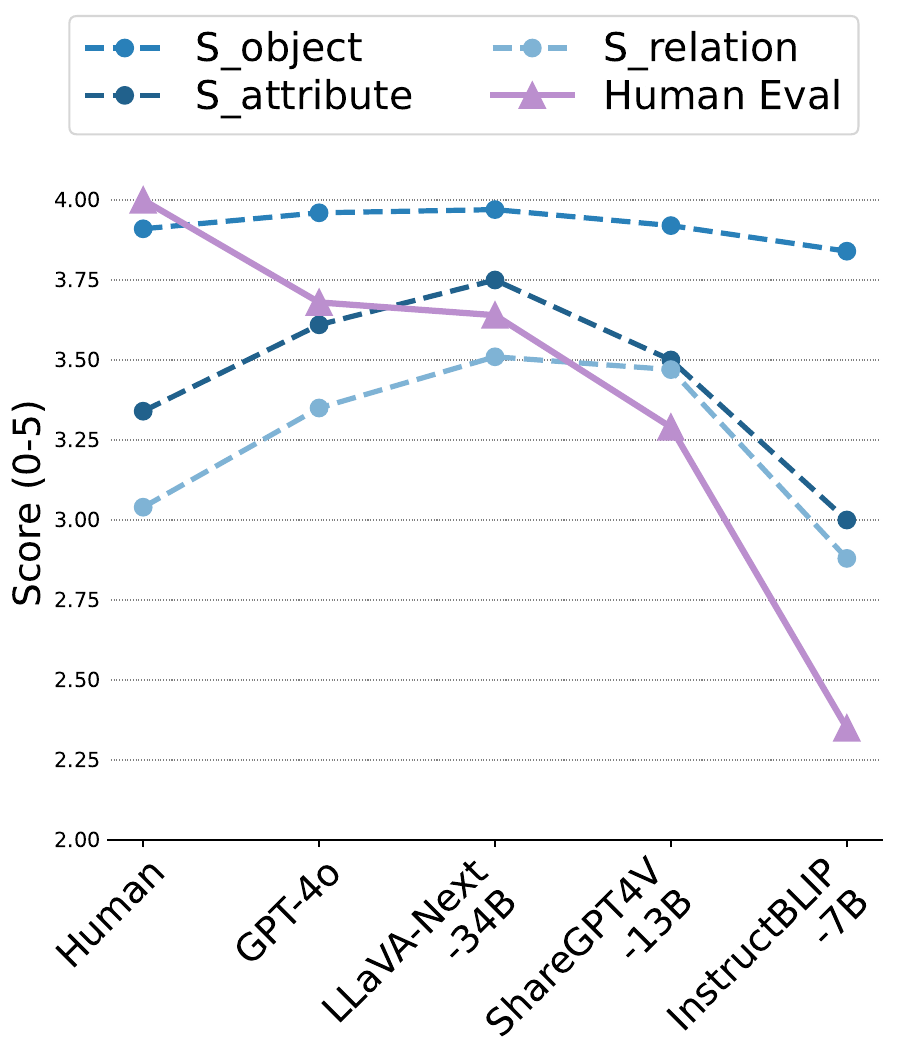}
     }
    \subfloat[\textbf{Our evaluation method}]{
         \centering
         \includegraphics[width=0.49\linewidth]{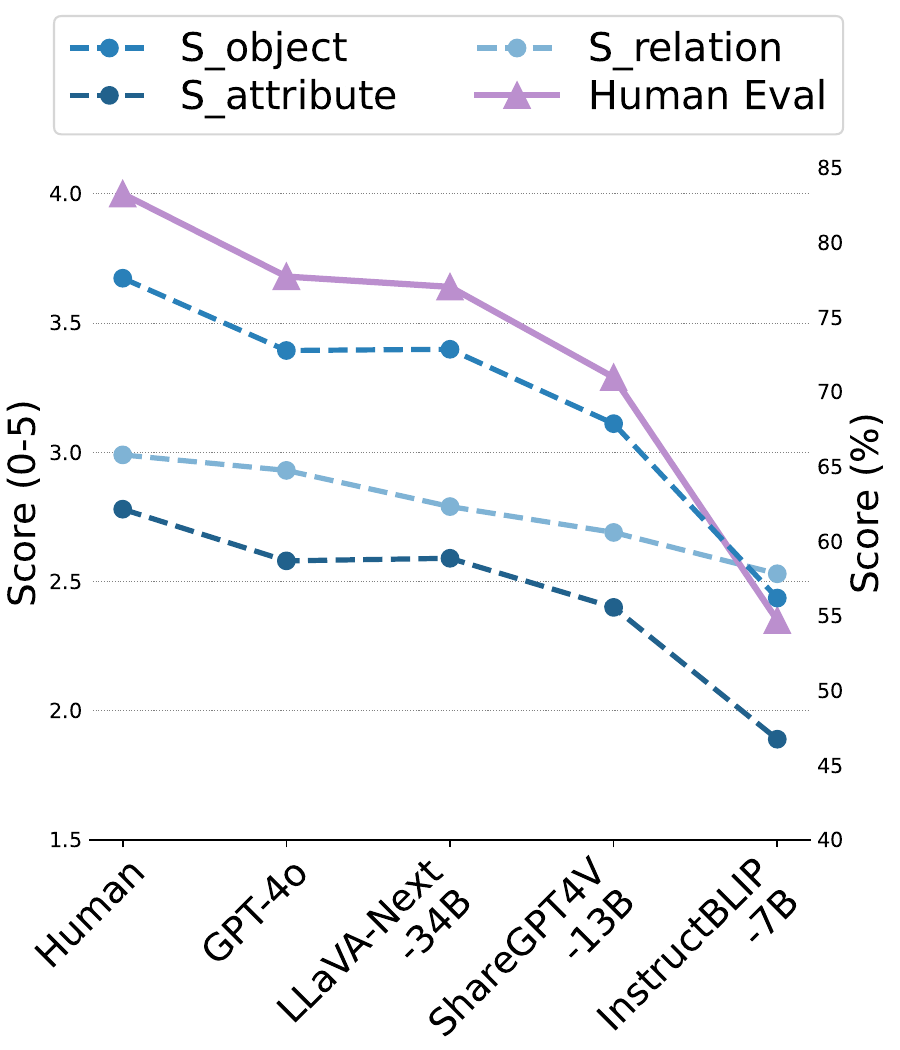}
    }
    \vspace{-1mm}
    \caption{Our method demonstrates a high consistency with human evaluation scores across all models, whereas the results directly produced by Llama3 exhibit conflicts. This indicates that the annotations of the directed scene graph facilitate more accurate matching at the levels of objects, attributes and relations during comprehensive caption assessment.}
    \label{supp_fig:consistency}
\end{figure}

We select the human performance, two top-performing
models in Tab.~\red{2} of the manuscript paper, GPT-4o and LLaVA-Next-34B, as well as a medium-performing model, ShareGPT4V13B, and a low-performing model, InstructBLIP7B, for comparative analysis. 
With the human assessment scores of the image captions
generated by both LVLMs and human, the metrics derived
from our method faithfully align with human evaluation
scores compared to those directly obtained from Llama3,
as shown in~\cref{supp_fig:consistency}. The high consistency with human
judgment emphasizes the credibility and practicality of our
CompreCap, proving its promising prospects in LVLMs
evaluation.

\subsection{Ability of Llama3 to Distinguish Caption Quality}
With the human annotation in the format of directed scene graph, we can apply Llama3~\cite{llama3} to assess the attribute descriptions bound to objects and directional relation descriptions between objects, instead of individual matching the words of objects, attributes and relations. To verify the ability of Llama3 to distinguish caption quality, we construct 50 pairs of good and bad captions by randomly shuffling object attributes and swapping the subject and object in relation descriptions, as illustrated in Fig.~\red{2} of the manuscript paper. We show some examples in~\cref{supp_fig:Llama3_error}. Then, we require Llama3~\cite{llama3} to score these captions on a scale from 0 to 5 using the prompt similar to that in  Fig.~\red{3} of the manuscript paper. We compare the scores with those obtained from the evaluation method of DetailCaps~\cite{detailcaption}, which isolately assess the words of objects, attributes and relations. The comparison in~\cref{supp_table:Llama3_error} reveals that Llama3~\cite{llama3} effectively discerns factual errors, such as those illustrated by the bad captions when the correct structure of scene graph is disrupted.

\subsection{Analysis on the influence of LLM evaluator}
\label{sec.llm}
\begin{figure}[t]
    \centering
    \includegraphics[width=\linewidth]{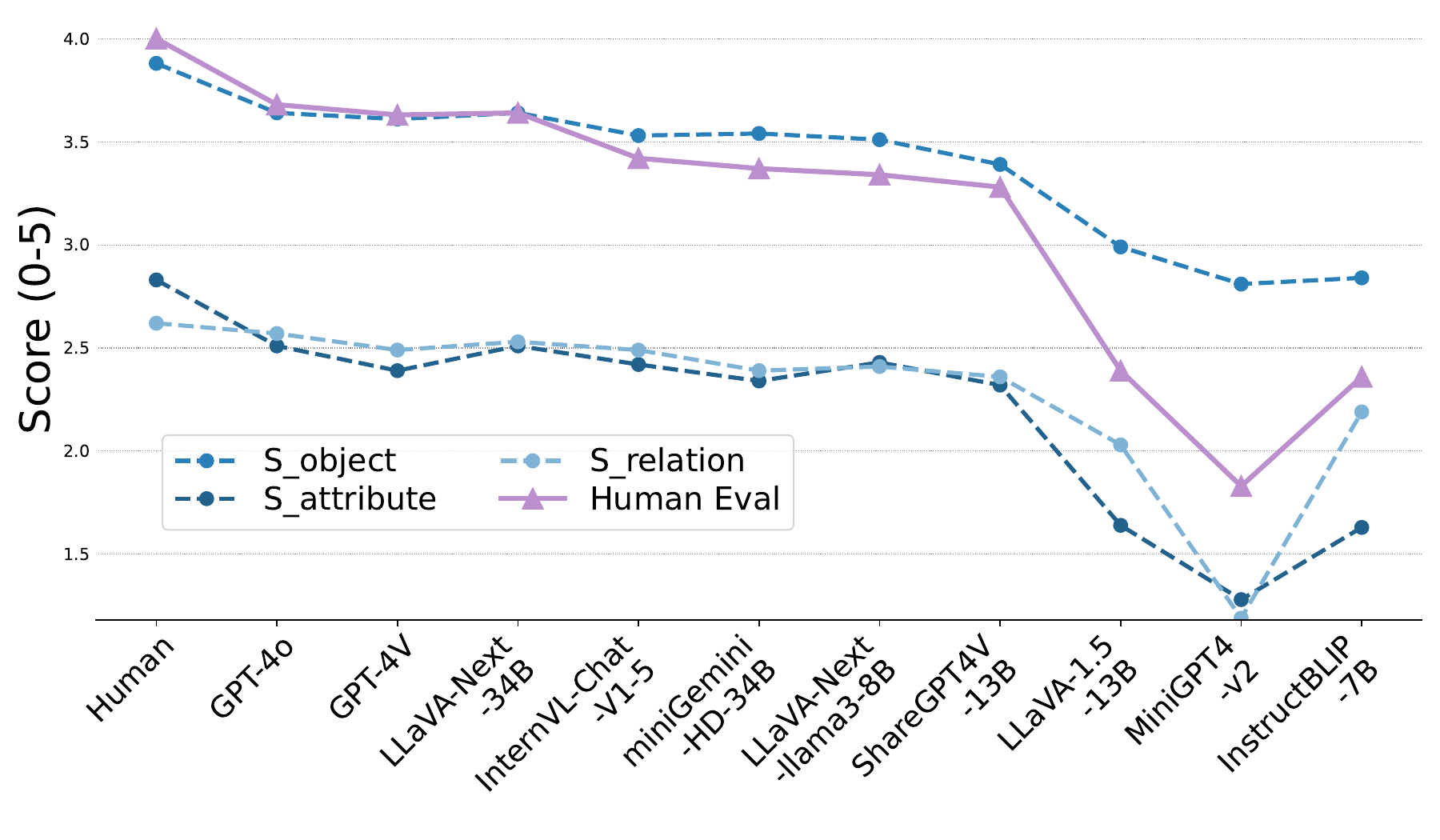}
    \captionsetup{skip=7pt}
    \captionof{figure}{The consistency of results using GPT4~\cite{gpt4} with human evaluation scores across 10 LVLMs and human performance. All our metrics obtained using GPT4~\cite{gpt4} align with human judgment.
    }
    \label{supp_fig:llm}
\vspace{-3mm}
\end{figure}

GPT4~\cite{gpt4} is known as the best close-source LLM. In this part, we employ GPT4 as 
the evaluator to assess the quality of detailed captions generated by 10 LVLMs and human. The prompt used for evaluation is the same as that in Fig.~\red{3} of the manuscript paper. The evaluation results presented in~\cref{supp_fig:llm} show that our metrics obtained by using GPT-4 also demonstrate a strong alignment with human evaluation scores. 
We choose LLama3 as the evaluator considering the advantages of non-API evaluation in terms of speed and stability.

\begin{table*}[t]
    \caption{Directly scoring the detailed captions generated by the 10 LVLMs with Llama3~\cite{llama3} on \method benchmark. The \textbf{best} results are highlighted in bold. The \underline{second} and \underline{\underline{third}} best results are highlighted in underline and double underline, respectively. The scores directly output by  Llama3~\cite{llama3} across various dimensions show a weak consistency with human evaluation scores.}
    \label{supp_tab:llama_longcap_main_whumanscoring}
    \centering
    \scriptsize
    \SetTblrInner{rowsep=1.5pt}      
    \SetTblrInner{colsep=2.pt}      
    \resizebox{\linewidth}{!}{
    \begin{tblr}{
        cells={halign=c,valign=m},   
        hline{1,2,13}={1.0pt},         
        hline{10,12}={},
        vline{2,3,7}={},         
        row{12}={c,gray!20},
    }
        Model &{ Caption\\ Length}&{ Object\\ Score} &{ Attribute\\ Score} &{ Relation\\ Score} &{ Overall\\ Score} & Human Evaluation \\

        InstructBLIP-7B~\cite{dai2024instructblip} & 69.93& 3.84\std{0.01}& 3.00\std{0.02}& 2.88\std{0.02}& 3.18\std{0.02} & 2.36\\
        MiniGPT4-v2~\cite{zhu2023minigpt} & 350.42& 2.26\std{0.01}& 1.67\std{0.02}& 1.32\std{0.02}& 1.66\std{0.02} & 1.83\\
        LLaVA-1.5-13B~\cite{liu2024visual} & 86.97& 3.79\std{0.00}& 2.89\std{0.01}& 2.73\std{0.01}& 3.07\std{0.01} & 2.39 \\
        ShareGPT4V-13B~\cite{chen2023sharegpt4v}& 155.91& 3.92\std{0.01}& 3.50\std{0.01}& \underline{\underline{3.47}}\std{0.02}& 3.68\std{0.01} & 3.28\\
        LLava-Next-llama3-8B~\cite{liu2024llava} &168.99& \underline{\underline{3.96}}\std{0.00}& 3.62\std{0.00}& \underline{3.50}\std{0.02}& \underline{\underline{3.76}}\std{0.01} & 3.34\\
        miniGemini-HD-34B~\cite{li2024mini}& 173.71& \underline{3.97}\std{0.00}& \textbf{3.76}\std{0.01}& \underline{3.50}\std{0.02}& \textbf{3.81}\std{0.01} & 3.37\\
        InternVL-Chat-V1-5~\cite{chen2023internvl}&  115.22& \textbf{3.98}\std{0.01}& 3.62\std{0.01}& 3.32\std{0.02}& 3.69\std{0.01} & 3.42\\
        LLava-Next-34B~\cite{liu2024llava}&  179.24& \underline{3.97}\std{0.01}& \underline{3.75}\std{0.01}& \textbf{3.51}\std{0.00}& \underline{3.79}\std{0.01} & 3.64 \\
        GPT-4V~\cite{yang2023dawn} & 202.06& \underline{\underline{3.96}}\std{0.00}& \underline{\underline{3.74}}\std{0.01}& 3.34\std{0.01}& 3.74\std{0.01} & 3.63 \\
        
        GPT-4o~\cite{gpt4o} &108.20& \underline{\underline{3.96}}\std{0.00}& 3.61\std{0.01}& 3.35\std{0.01}& 3.67\std{0.00} & 3.68\\

        Human & 133.61& 3.91\std{0.00}& 3.34\std{0.01}& 3.04\std{0.01}& 3.40\std{0.01} & 4.0\\

        \end{tblr}
        }
\end{table*}

It is important to note that, to fully leverage the discriminative power of Llama3~\cite{llama3}, it is necessary to decouple the detailed captions and match them with annotations at the levels of object, attribute, and relation. Otherwise, as highlighted in~\cref{supp_sec:compare_direct_llama}, directly scoring the detailed captions with dense text using Llama3~\cite{llama3} fails to yield solid evaluation results on \method.

\begin{figure}[t]
    \centering
\captionsetup{skip=3pt}
    \caption{We construct the bad captions through shuffling the attributes and swapping the subject and object in relation descriptions. The correct and incorrect components compared with the reference caption are marked in {\color{cellgreen}green} and {\color{red}red}, respectively.}
    \label{supp_fig:Llama3_error}

    \begin{minipage}{1.0\columnwidth}
    \centering
    \begin{tcolorbox} 
        \centering
        \small
         \hspace{-6mm}
    \begin{minipage}{1.0\columnwidth}
    \textbf{Reference:} A blue refrigerator is next to a white cabinet. \\
    \textbf{Good Caption:} There are a {\color{cellgreen}blue} refrigerator and a {\color{cellgreen}white} cabinet. \\
    \textbf{Bad Caption:} There are a {\color{red}white} refrigerator and a {\color{red}blue} cabinet. \\
    --------------------------------------------------------------------
    \textbf{Reference:} A woman is in front of the panda. \\
    \textbf{Good Caption:} There is a {\color{cellgreen}woman} in front of the {\color{cellgreen}panda}. \\
    \textbf{Bad Caption:} There is a {\color{red}panda} in front of the {\color{red}woman}.
    \end{minipage}
    \end{tcolorbox}
    \end{minipage}
\end{figure}

\begin{figure}[t]
    \centering
    \includegraphics[width=0.99\linewidth]{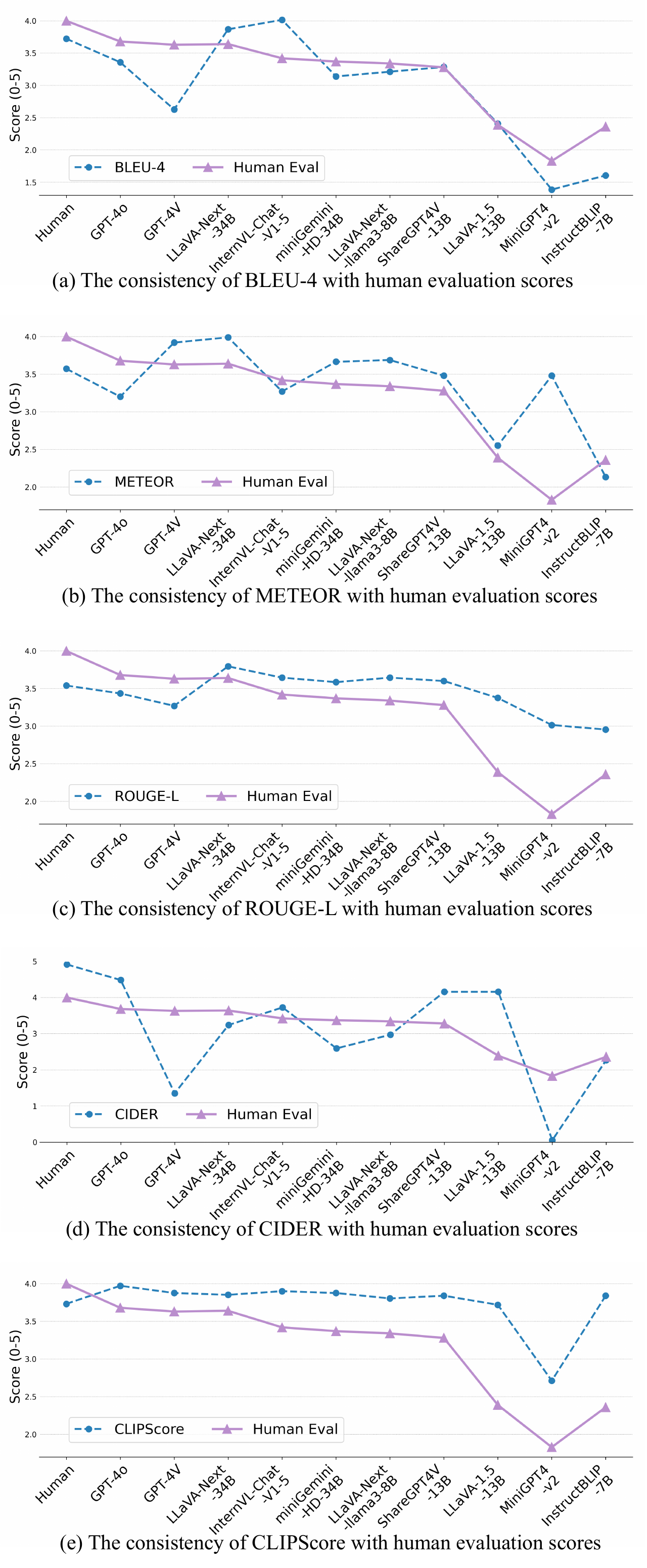}
    \vspace{-3mm}
    \captionof{figure}{The consistency of traditional caption metrics with human evaluation scores across 10 LVLMs and human performance. All traditional caption metrics fail to align with human judgment. The scores of all traditional metrics are linearly scaled to 0-5. 
    }
    \label{supp_fig:traditional_metrics}
\vspace{-3mm}
\end{figure}

\subsection{Analysis on the effect of parsers}
\begin{figure}[t]
    \centering
    \includegraphics[width=\linewidth]{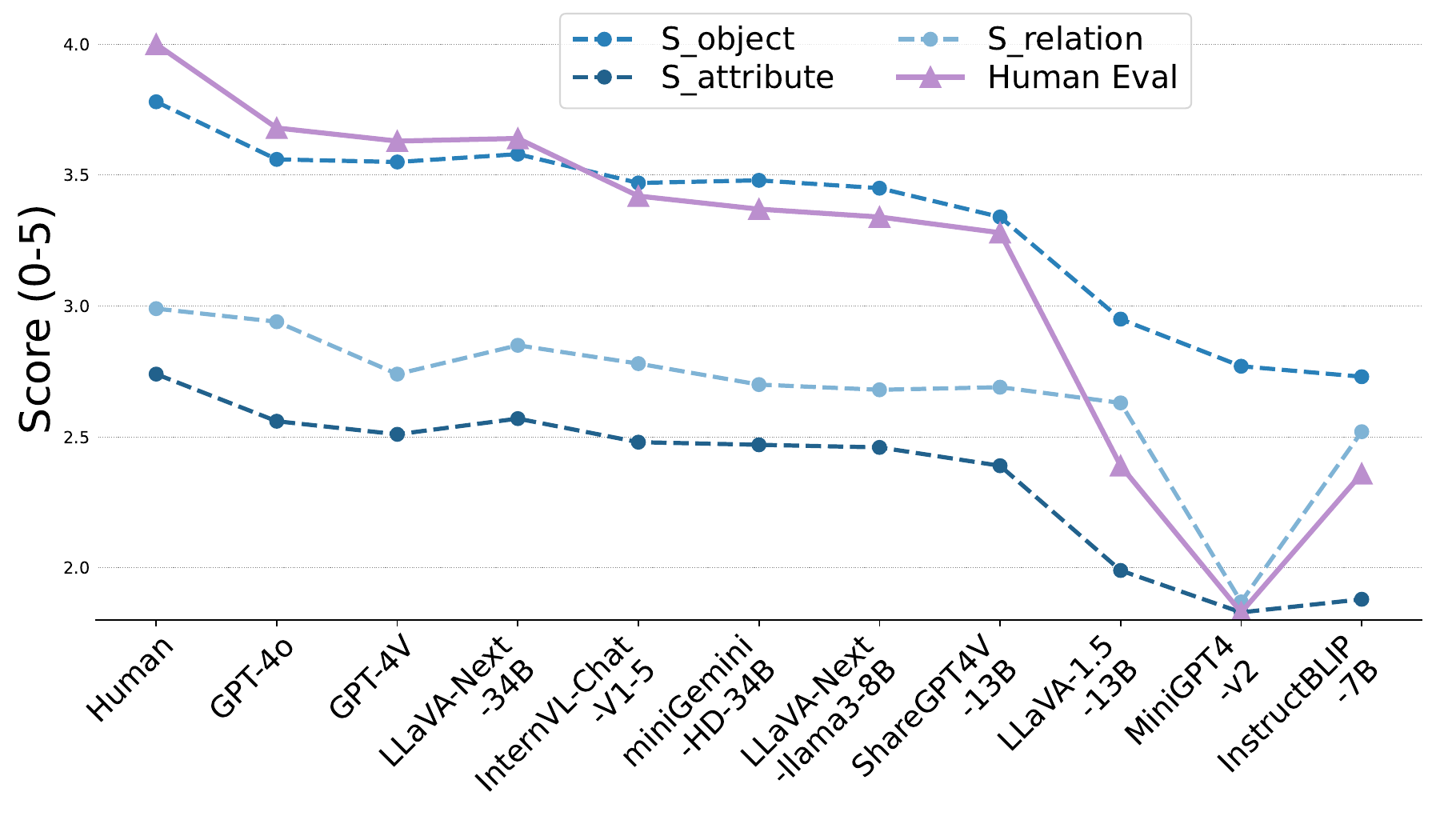}
    \captionsetup{skip=5pt}
    \captionof{figure}{The consistency of results based on NLTK~\cite{honnibal2020industrial} with human evaluation scores across 10 LVLMs and human performance. All our metrics obtained based on NLTK~\cite{honnibal2020industrial} align with human judgment.
    }
    \label{supp_fig:parser}
\end{figure}

Our evaluation method applies the spaCy~\cite{honnibal2020industrial} parser to parse nouns and decouple the generated captions into a hierarchical structure. We present the consistency of results using another well-known parser, \ie, NLTK~\cite{loper2002nltk}, with human evaluation scores in~\cref{supp_fig:parser}. The high consistency shows that our metrics are not sensitive to parsers.

\subsection{Results of traditional caption metrics on CompreCap}
We first evaluate 10 LVLMs and human performance with traditional caption metrics, including BLEU-4~\cite{papineni2002bleu}, METEOR~\cite{banerjee2005meteor}, ROUGE-L~\cite{lin2004rouge}, CIDER~\cite{vedantam2015cider}, and CLIPScore~\cite{hessel2021clipscore}, on our \method dataset, then show the consistencies of these traditional metrics with human evaluation scores in~\cref{supp_fig:traditional_metrics}. The scores of captions generated by human do not outperform all LVLMs on all traditional metrics. And all traditional caption metrics fail to align with human judgment. The results show that existing traditional metrics cannot reasonably evaluate comprehensive image captions consisting of dense text.

\begin{table}[btp!]
\centering
\setlength{\tabcolsep}{1.5mm}
{
\captionsetup{skip=5pt}
\caption{The score gaps between good and bad captions. The comparison reveals that Llama3~\cite{llama3} effectively distinguishes the quality difference between good and bad captions. `Good' and `Bad' denote `good captions' and `bad captions', respectively. We conducted three tests repeatedly, and the mean along standard deviation are reported.}
\label{supp_table:Llama3_error}
\begin{tabular}{cc|cc}
\toprule 
\multicolumn{2}{c|}{DetailCaps (F1-score)}    & \multicolumn{2}{c}{Llama3 (\text{0}$\sim$\text{5})}   \\      
\cline{1-4}
\small{Good} & \small{Bad} & \small{Good} & \small{Bad} \\ 
83.15\std{0.00} & 82.83\std{0.00} & 4.75\std{0.00} & 1.98\std{0.02} \\
\bottomrule
\end{tabular}}
\end{table}

\subsection{Analysis on the stability of generated detailed caption}
In Tab.~\red{2} of the manuscript, we include error bars in the evaluation results to certify the reliability and consistency of our evaluation methodology.
However, LVLMs can generate different detailed captions even we use the same prompt for guidance.
To investigate the fluctuations in the quality of detailed captions, we repeatedly utilize GPT-4o~\cite{gpt4o} and LLava-Next-34B~\cite{liu2024llava} to generate detailed captions three times and evaluate their quality. The average performances along the error bars are reported in~\cref{supp_tab:diff_seed_longcap}. Although the averaged length of the generated detailed captions differs each time, the evaluation metric scores remain relatively consistent.

\begin{table*}
\captionsetup{skip=5pt}
    \caption{Investigate the fluctuations in the detailed captions generated by GPT-4o and LLava-Next-34B.}
    \label{supp_tab:diff_seed_longcap}
    \centering
    \scriptsize
    \SetTblrInner{rowsep=1.0pt}      
    \SetTblrInner{colsep=6.0pt}      
    \resizebox{\linewidth}{!}{
    \begin{tblr}{
        cells={halign=c,valign=m},   
        hline{1,2,4}={1.0pt},         
        vline{2,4,5}={},         
    }
        Model & Visual Encoder & LLM  &{ Caption\\ Length}&\( S_{\text{object}} (\%)\uparrow\) &\( S_{\text{attribute}} \uparrow \) &\( S_{\text{relation}} \uparrow\)& \( \text{S-Cov.} (\%)\uparrow\)  \\
        LLava-Next-34B~\cite{liu2024llava}& CLIP-ViT-L/14 & Yi-34B & 179.87\std{0.67}& 72.46\std{0.30}& 2.57\std{0.01}& 2.79\std{0.01}& 58.47\std{0.41}\\
        GPT4o~\cite{yang2023dawn}& - & - &108.06\std{0.21}& 72.79\std{0.12}& 2.58\std{0.00}& 2.92\std{0.02}& 57.62\std{0.18}  \\

        \end{tblr}
        }
\end{table*}


\begin{table*}[t]
    \caption{Analysis on the effect of prompts used for detailed caption generation. {Prompt \#1}: `Please describe the image in detail.'. {Prompt \#2}: `Please describe the image in detail, focusing on the visible objects.' {Prompt \#3}: `Please describe the image in detail, focusing on the visible objects and the relationships among these objects.' }
    \label{supp_tab:longcap_prompt}
    \centering
    \scriptsize
    \SetTblrInner{rowsep=1.2pt}      
    \SetTblrInner{colsep=8.0pt}      
    \resizebox{\linewidth}{!}{
    \begin{tblr}{
        cells={halign=c,valign=m},   
        hline{1,2,8}={1.0pt},         
        hline{5}={},
        vline{2,3,4}={},         
        cell{2,5}{1}={r=3}{},
    }
        Model & Prompt & Caption Length & \( S_{\text{object}}(\%) \uparrow\) &\( S_{\text{attribute}} \uparrow \) &\( S_{\text{relation}} \uparrow\) & \( \text{S-Cov}(\%) \uparrow\)  \\

        GPT4o~\cite{hurst2024gpt} & {\#1} & 90.46& 70.22& 2.48\std{0.00}& 2.89\std{0.00}& 56.74\std{0.18}\\
         & {\#2} & 91.05& 71.38& 2.54\std{0.00}& 2.92\std{0.01}& 57.20\std{0.07} \\
         & {\#3} & 108.20& 72.78& 2.58\std{0.00}& 2.93\std{0.00}& 57.54\std{0.23} \\

         LLava-Next-34B~\cite{liu2024llava} & {\#1} & 153.38& 69.38& 2.45\std{0.00}& 2.71\std{0.00}& 56.96\std{0.05}\\
         & {\#2} & 154.22& 70.90& 2.53\std{0.01}& 2.78\std{0.00}& 57.62\std{0.11} \\
         & {\#3} & 179.24& 72.86& 2.59\std{0.00}& 2.79\std{0.00}& 58.49\std{0.15}\\

        \end{tblr}
        }
\end{table*}

\subsection{Analysis of prompts for caption generation}
We analyze the effects of prompts used for caption generation with three different prompts in~\cref{supp_tab:longcap_prompt}. The compared results between prompts \#1 and \#2 indicate that emphasizing the descriptions of objects can enhance the quality of generated captions across all metrics. And prompt \#3 which highlights both objects and relations further improves the comprehensiveness of generated detailed captions. Overall, more explicit and detailed prompts contribute to the generation of higher-quality captions by LVLMs.

\subsection{Analysis on the failure cases in the fine-grained object VQA}
\begin{figure*}
\centering 
\subfloat[InstructBLIP-7B]{
     \centering
     \includegraphics[width=0.31\linewidth]{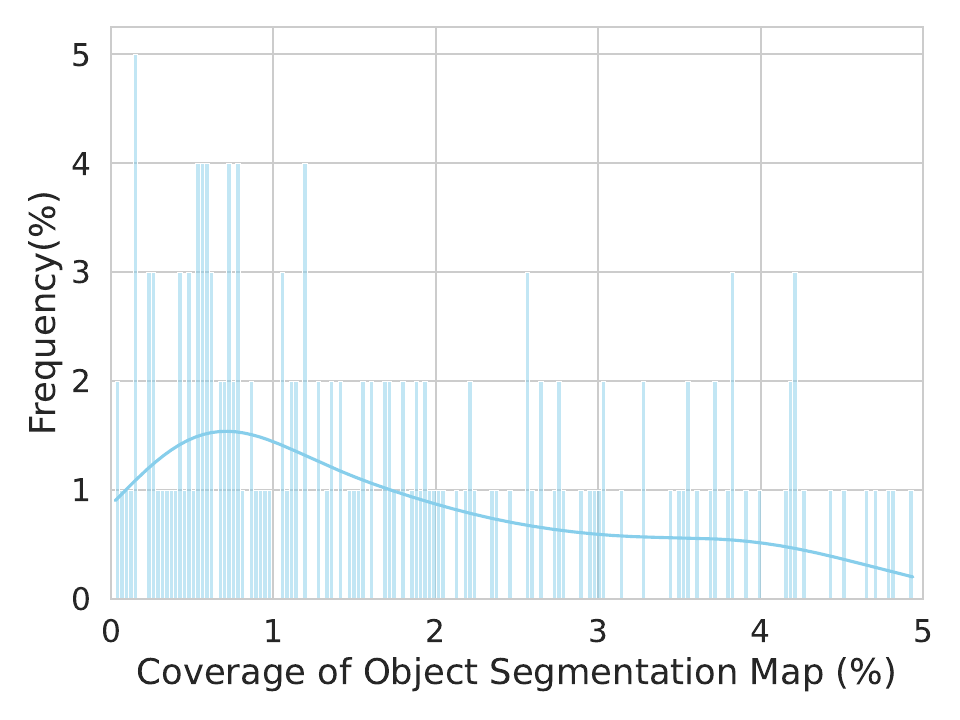}
 }
 \hfill
\subfloat[MiniGPT4-v2]{
     \centering
     \includegraphics[width=0.31\linewidth]{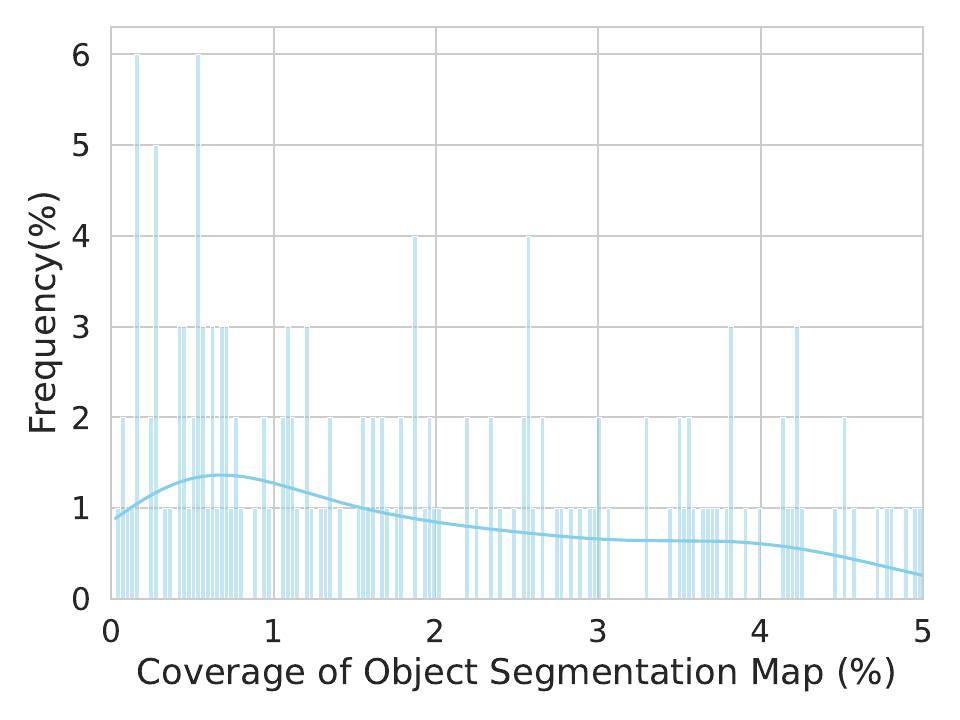}
}
 \hfill
\subfloat[ShareGPT4V-13B]{
     \centering
     \includegraphics[width=0.31\linewidth]{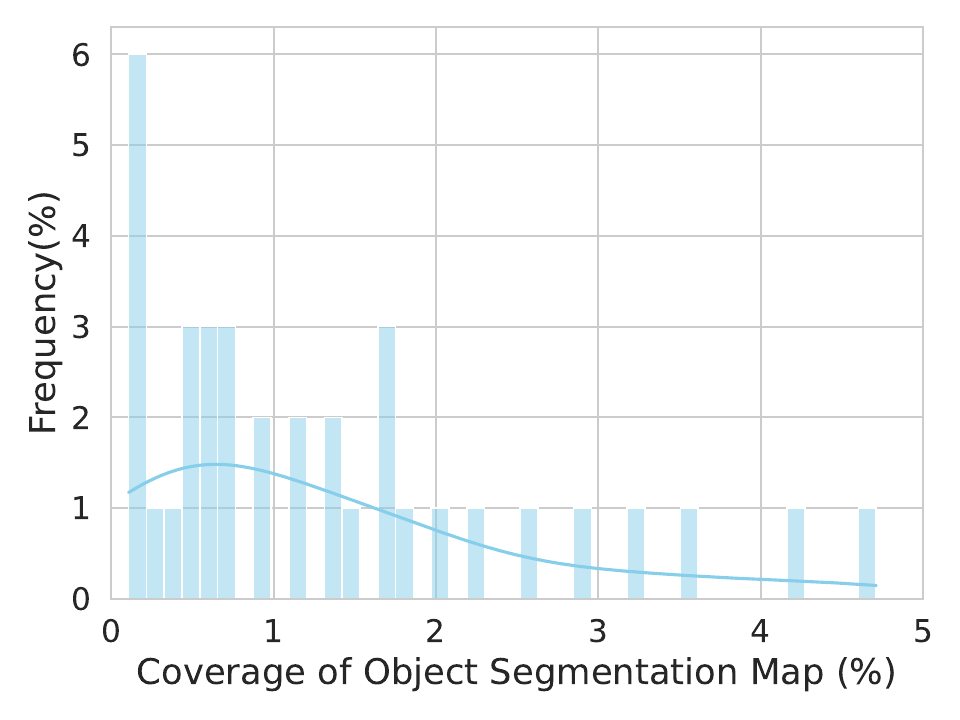}
}
 \hfill
\subfloat[LLaVA-1.5-13B]{
     \centering
     \includegraphics[width=0.31\linewidth]{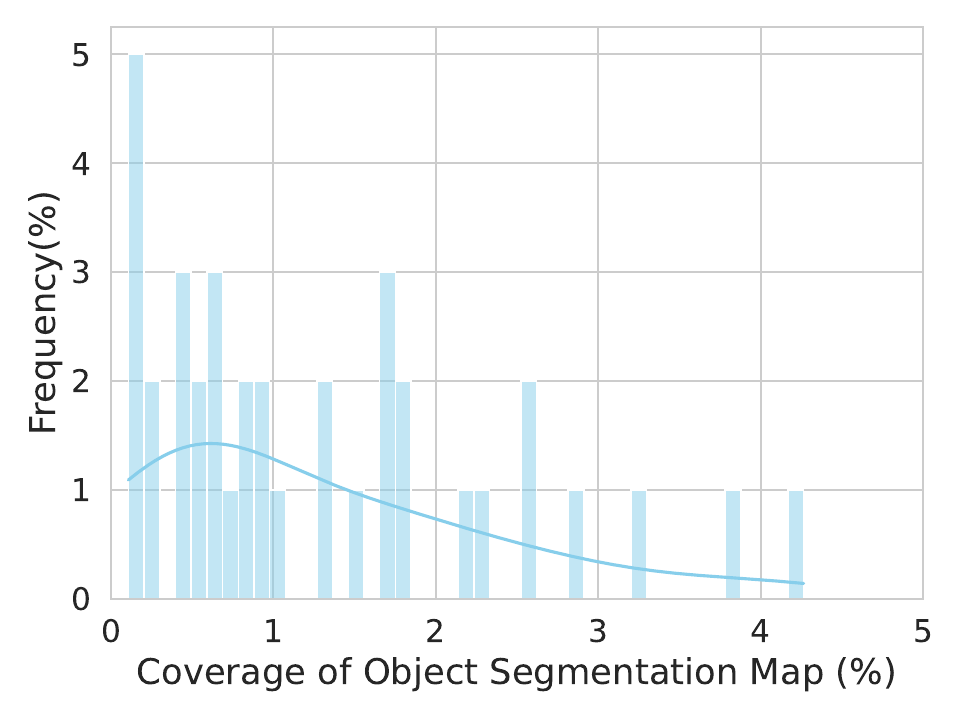}
 }
 \hfill
\subfloat[miniGemini-34B-HD]{
     \centering
     \includegraphics[width=0.31\linewidth]{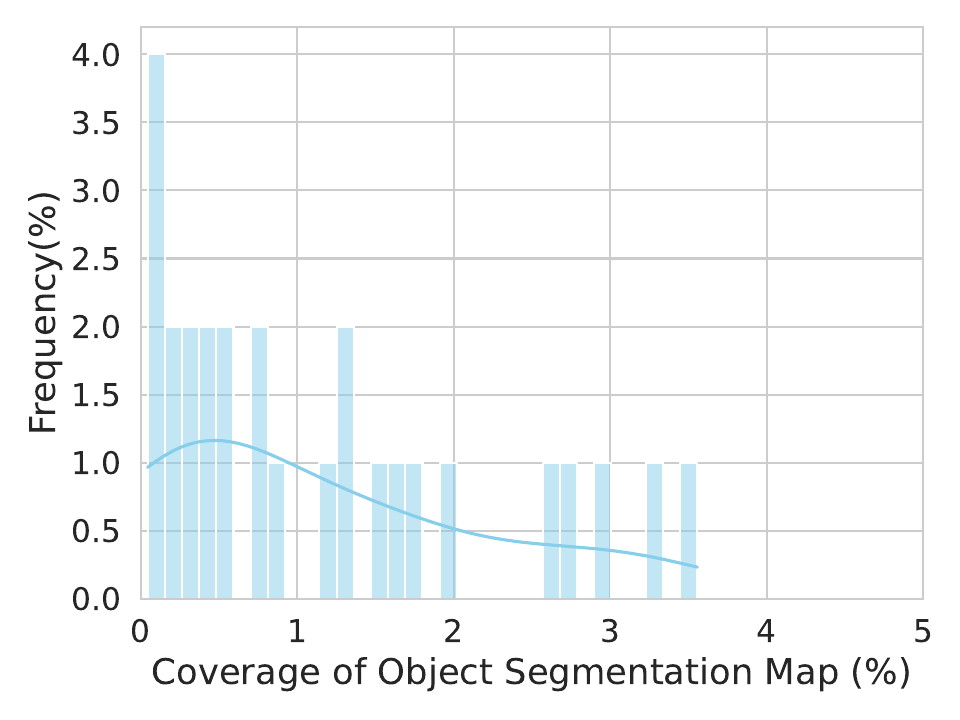}
}
 \hfill
\subfloat[LLava-Next-llama3-8B]{
     \centering
     \includegraphics[width=0.31\linewidth]{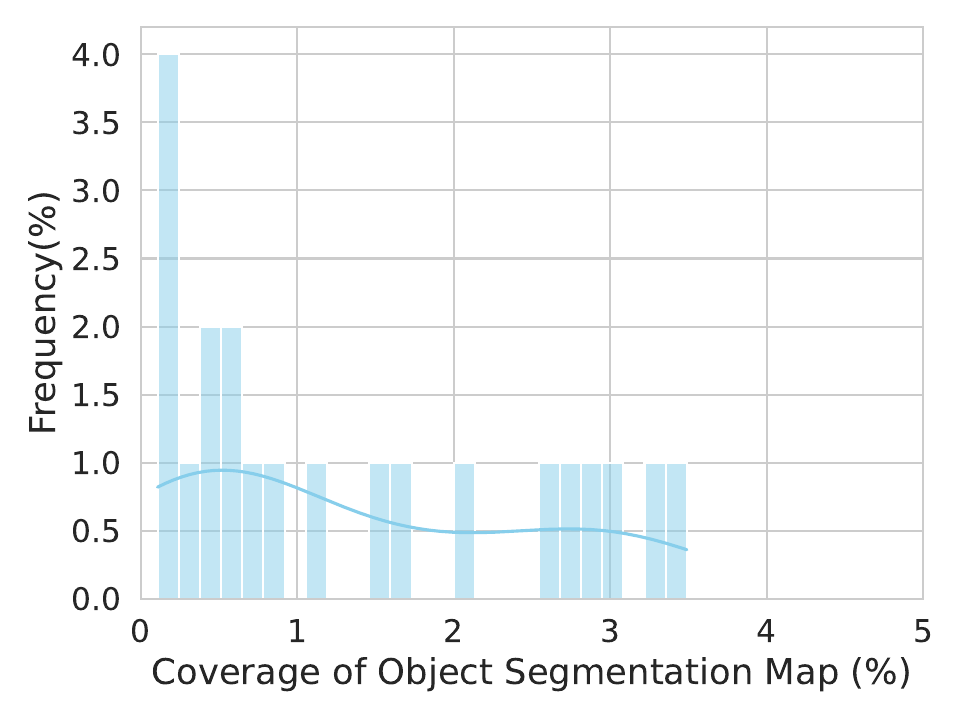}
}
 \hfill
\subfloat[LLava-Next-34B]{
     \centering
     \includegraphics[width=0.31\linewidth]{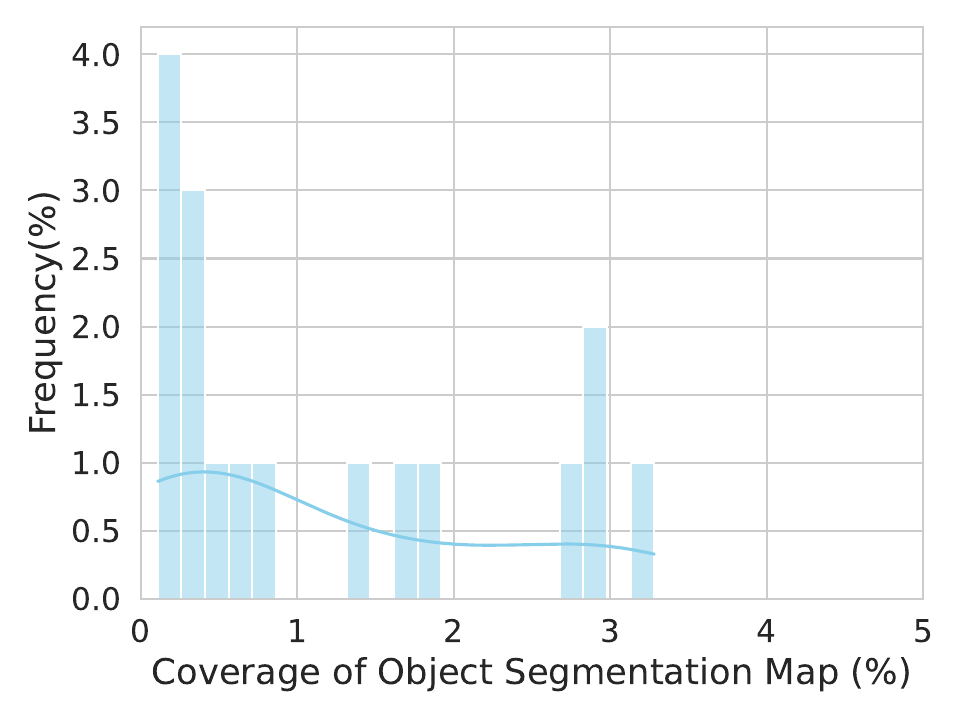}
 }
 \hfill
\subfloat[InternVL-Chat-V1-5]{
     \centering
     \includegraphics[width=0.31\linewidth]{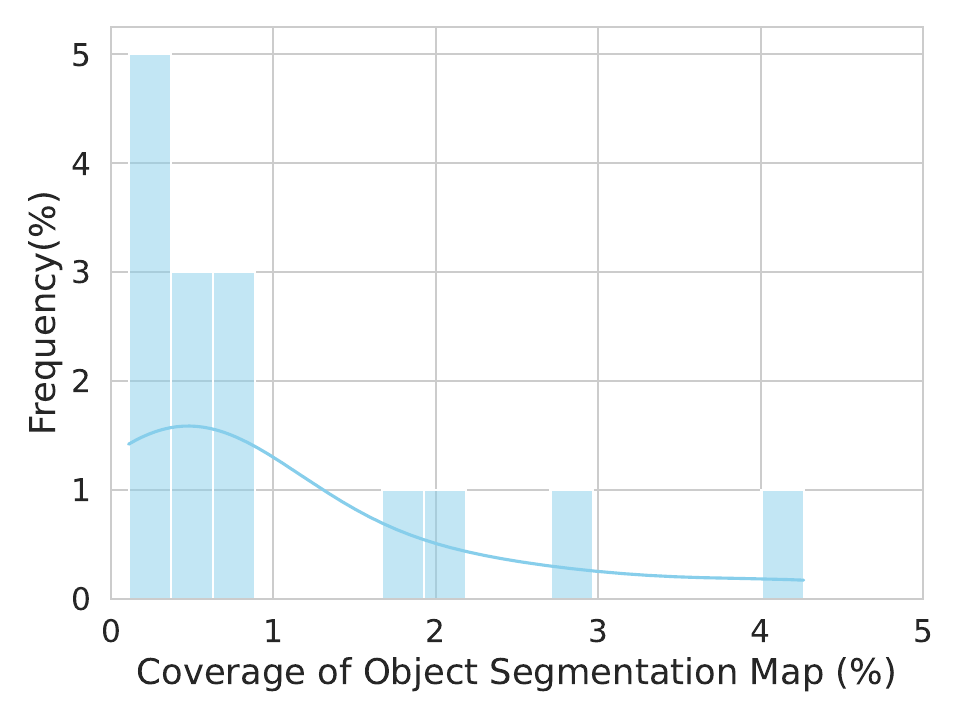}
}
 \hfill
\subfloat[GPT-4o]{
     \centering
     \includegraphics[width=0.31\linewidth]{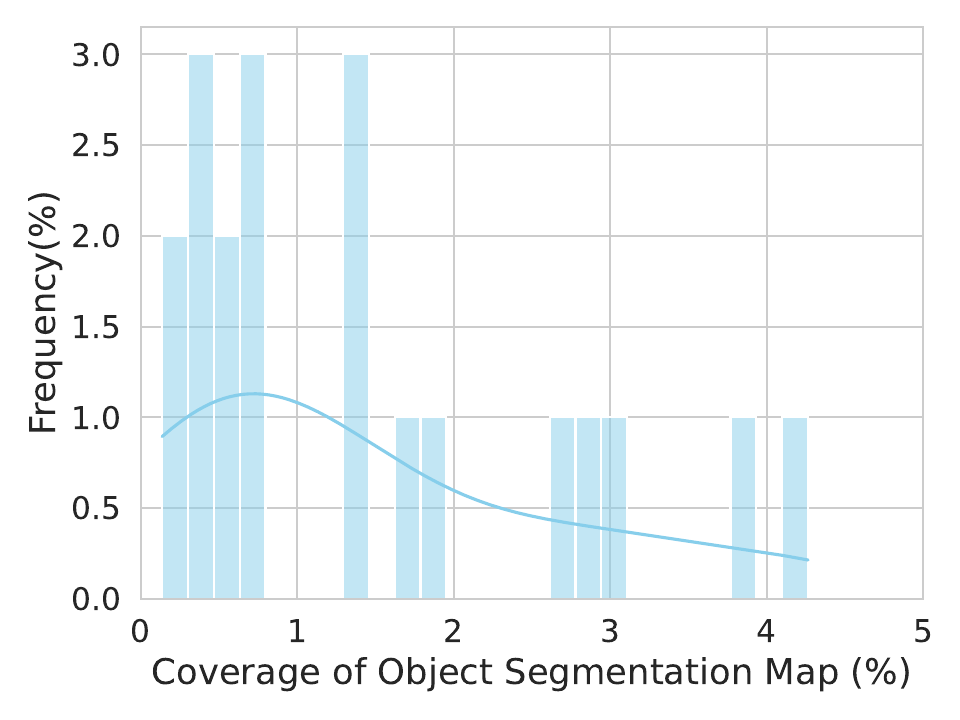}
}
\caption{We present the error objects' segmentation map coverage in CompreQA-for-Caption of 9 LVLMs. The distribution shows that LVLMs tend to choose inaccurate captions for tiny objects whose segmentation map coverage \textless 2\%.}
\label{fig:sata_qa}
\end{figure*}

We analyze the segmentation map coverage distributions of 9 LVLMs' error objects in CompreQA-for-Caption and show the distributions in \cref{fig:sata_qa}. All 9 LVLMs tend to choose inaccurate captions for tiny objects whose segmentation map coverage \textless 2\%, indicating that the smaller the object, the greater the challenge for LVLMs to accurately describe it.

\begin{figure*}
    \centering
    \includegraphics[width=0.98\textwidth]{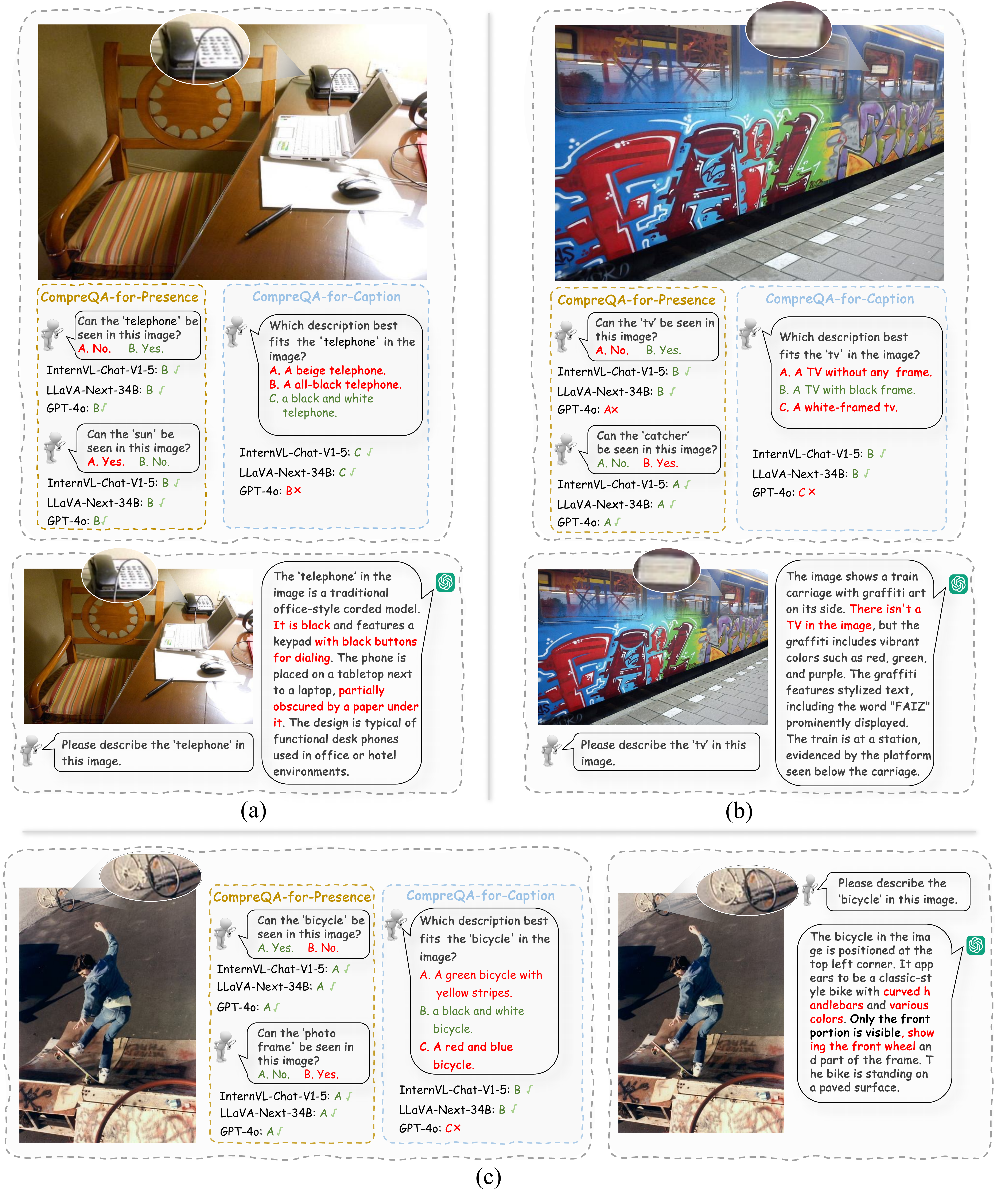}
\captionsetup{skip=5pt}
    \caption{The answer is denoted in green, while error options are in red. The input images do not include the magnification effect, which is solely applied for clearer display purposes. We also ask GPT-4o~\cite{gpt4o} to describe the tiny objects and label incorrect descriptions with red. The examples show that GPT-4o~\cite{gpt4o} fails to comprehend the visual content of tiny objects.}
    \label{supp_fig:comparison_qa2}
\end{figure*}

We present comparison examples mong LLava-Next-34B~\cite{liu2024llava}, InternVL-Chat-V1-5~\cite{chen2023internvl} and GPT-4o~\cite{gpt4o} on CompreQA-for-Presence and CompreQA-for-Caption in~\cref{supp_fig:comparison_qa2} and show cases where GPT-4o~\cite{gpt4o} fails in. We observe GPT-4o~\cite{gpt4o} fails to accurately understand fine-grained objects in the background (\eg, the `bicycle'). Then, we directly ask GPT-4o~\cite{gpt4o} to describe the tiny objects which are misinterpreted in CompreQA. The illustration in~\cref{supp_fig:comparison_qa2} shows that the descriptions generated by GPT-4o~\cite{gpt4o} conflict the visual content, as it overlooks the `tv' within the carriage and mischaracterizes the attributes of the `telephone' and `bicycle'.

\begin{table}[!ht]
\captionsetup{skip=5pt}
    \caption{Comparison with similar benchmarks. Visual-Genome includes VG-Attributes and VG-Relations.}
    \centering\footnotesize
    \setlength{\tabcolsep}{5.0pt}
    \begin{tabular}{l|c|c|c|c|c}
    \hline  
     & Length & Mask & Object  & Attribute & Relation \\
    \hline
    COCO-Stuff & 10 & \checkmark & \checkmark & - & - \\
    COCO-OOD &  10 & \checkmark & \checkmark & - & - \\
    SugarCrepe & 11 & - & \checkmark & - & - \\
    Visual-Genome & 92 & - & \checkmark & \checkmark  &\checkmark\\
    \rowcolor{lightgray!44} 
    \method &\textbf{172} & \checkmark & \checkmark & \checkmark & \checkmark \\
    \hline
    \end{tabular}
    \label{supp_label:R1}
\end{table}

\section{
Comparison with Similar Benchmarks}

Here we compare several related benchmarks in~\cref{supp_label:R1}, especially  Visual-Genome (VG)~\cite{krishna2017visual} and other COCO-based benchmarks. COCO-Stuff/OOD~\cite{hsieh2023sugarcrepe,caesar2018coco} have object and mask annotations, while SugarCrepe generates various forms of hard negatives based on COCO's short captions (averaged length is about 10 words). 
VG has object, attribute, and relation annotation with a bounding box, which has about 92 lengthy captions per image. Our CompreCap has a more detailed caption (about 172) and more dense mask annotation than VG, where the bounding box coverage has about 99.23\% \textit{vs.} 85.26\% of VG. For comprehensive image captioning, a dataset with wider coverage and a more detailed description is important. Moreover, based on our dataset, we also propose an evaluation metric for comprehensive image captioning.

They all lack annotations of object-bound attributes and directional relations. VG is the closest to our dataset, but its relatively brief annotations limit the diversity of attribute and relation descriptions, and its lower coverage cannot ensure the inclusion of all main objects in images, hindering the comprehensive evaluation of detailed captions. Moreover, VG does not propose a method to assess coherent long image captions. Instead, it makes the model output captions for different image regions sequentially and still uses traditional n-gram evaluation metrics.

The characteristics of \method are highlighted as follows:
\textbf{1. Comprehensive annotation:} \method has object, attribute, and relation annotations with pixel-level masks;
\textbf{2. More detailed caption:} \method has more detailed human-annotated captions, with an average of 172 words \textit{vs.} VG's 92;
\textbf{3. Wider coverage:} The pixel coverage of \method is 95.83\%. Its bounding box coverage is {99.23\% \textit{vs.} 85.26\%} of VG;
\textbf{4. Evaluation method for long caption of image:} Compared to other benchmarks, \method proposes a method to comprehensively evaluate the visual context in detailed captions.

\section{Data Comparison with MSCOCO}
\label{supp_sec:compare_w_coco}

\begin{table*}
    \caption{The data statistics of \method and a subset of MSCOCO panoptic segmentation validation dataset (MSCOCO$_{\text{sub}}$). MSCOCO$_{\text{sub}}$ contains the same images as \method. }
    \label{supp_tab:data_statistics}
    \centering
    \scriptsize
    \SetTblrInner{rowsep=2.0pt}      
    \SetTblrInner{colsep=3.0pt}      
    \resizebox{\linewidth}{!}{
    \begin{tblr}{
        cells={halign=c,valign=m},   
        hline{1,4}={1.0pt},         
        hline{2}={},         
    }
        Dataset & \# Images & \# Objects Categories & {\# Objects per Image} & {\# Relations per Image} & {Averaged Text Length \\ per Image} & \# Q/As \\
        
         MSCOCO$_{\text{sub}}$~\cite{lin2014microsoft} & 560& 131 & 5.46 & - & 10 & -   \\
         \method & 560& 412 & 6.26& 8.84 & 172& 846  \\
          
        \end{tblr}
        }

\end{table*}

\begin{figure*}
    \centering
    \includegraphics[width=0.95\textwidth]{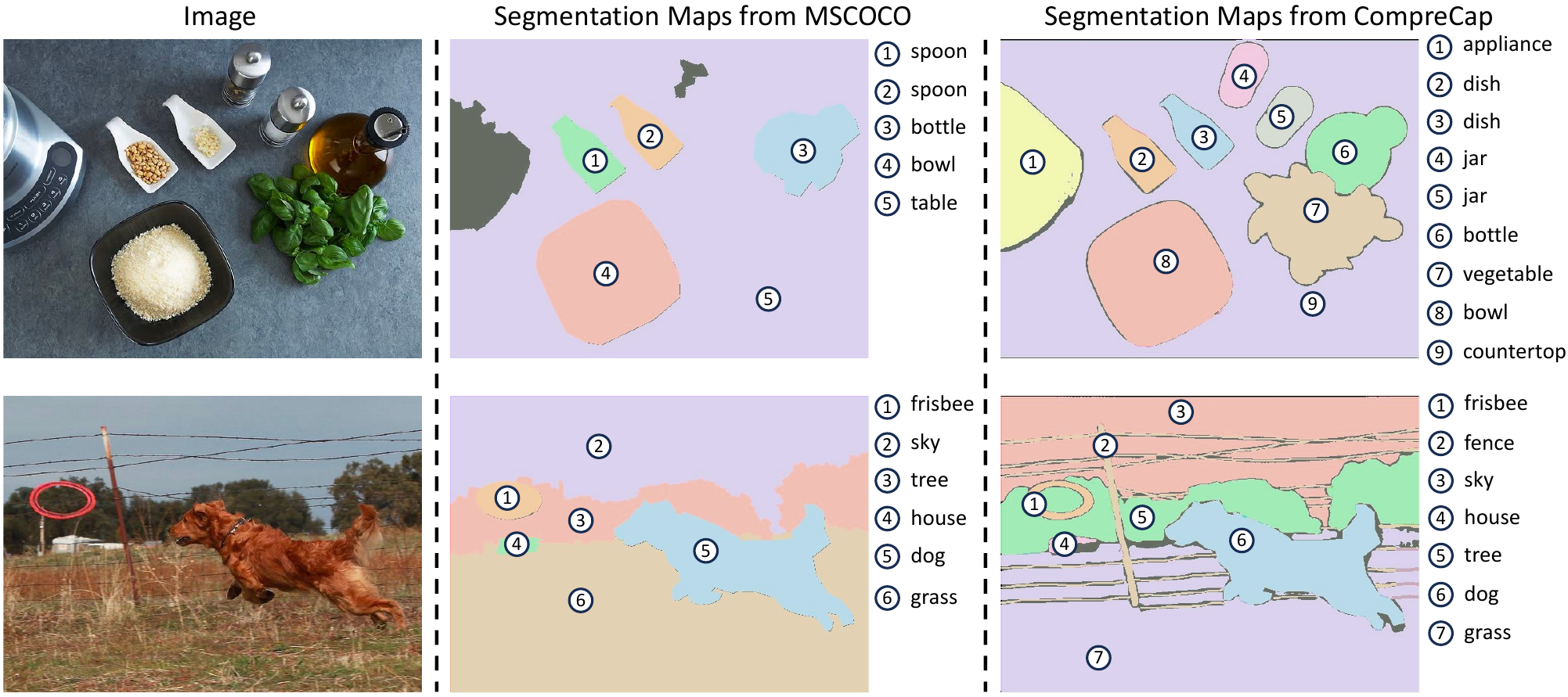}
    \caption{We refine the object and segmentation annotations based on MSCOCO panoptic segmentation dataset. Concretely, we add or re-annotate the class labels and more precise segmentation maps for the common objects within images, \eg `jar' in the first raw, and `fence' in the second raw. Besides, we improve the accuracy of pixel-level annotation, \eg the segmentation map of `frisbee' in the second raw.}
    \label{supp_fig:seg_compare}
\end{figure*}
In Sec.~\red{3.1} of the manuscript paper, we describe the process of sampling 560 images from the MSCOCO~\cite{lin2014microsoft} Panoptic Segmentation dataset.
MSCOCO Panoptic Segmentation dataset includes instance segmentation annotations for objects within images. Each segmentation map corresponds to an object in the image. 
However, we observed that segmentation maps for certain visible objects are missing. These missing objects are beyond the common objects listed in MSCOCO Panoptic Segmentation dataset. 
Additionally, we noticed that some of the segmentation maps for common objects overlapped with other visible objects in the image.
Thus, we re-annotate the 560 images.
Concretely, we first collect a common categories vocabulary from several well-known datasets including RAM~\cite{zhang2024recognize}, COCO~\cite{lin2014microsoft}, OpenImagesV4~\cite{kuznetsova2020open}, and Object365~\cite{shao2019objects365}. Then, we re-annotate category labels and more precise segmentation maps for common objects in images within the vocabulary. The examples of re-annotated segmentation maps are demonstrated in~\cref{supp_fig:seg_compare}.
Moreover, we present a comparison on the data statistics of the \method dataset with a subset of the MSCOCO Panoptic Segmentation dataset (MSCOCO$_{\text{sub}}$) in~\cref{supp_tab:data_statistics}. Both datasets contain the same set of images, but \method dataset includes additional annotations such as more types of objects, longer texts, and relations between objects.

\begin{table*}
    \vspace{-3mm}
    \caption{Comparison of \method with long caption datasets.}
    \label{supp_tab:data_composition}
    \centering
    \scriptsize
    \SetTblrInner{rowsep=0.5pt}      
    \SetTblrInner{colsep=10.0pt}      
    \resizebox{\linewidth}{!}{
    \begin{tblr}{
        cells={halign=c,valign=m},   
        hline{1,2,6}={1.0pt},         
        hline{5}={},         
    }
        Dataset  & {Averaged\\ Text Length } & Object & Segmentation Map & Attribute & Relation & Q/A & Answer Type \\
        
         DOCCI~\cite{onoe2024docci}  & 136 & - & - & - & - & - & - \\
         IIW~\cite{garg2024imageinwords}  & 217 & - & - & - & - & -& - \\
         DCI~\cite{DCI}  & 148  & \checkmark & \checkmark & \checkmark &- & - & - \\
         \method & 172 & \checkmark & \checkmark  & \checkmark& \checkmark& \checkmark & A/B/C\\
        \end{tblr}
        }
\end{table*}

\section{Data Samples from \method dataset}
\begin{figure*}
    \centering
    \includegraphics[width=0.99\textwidth]{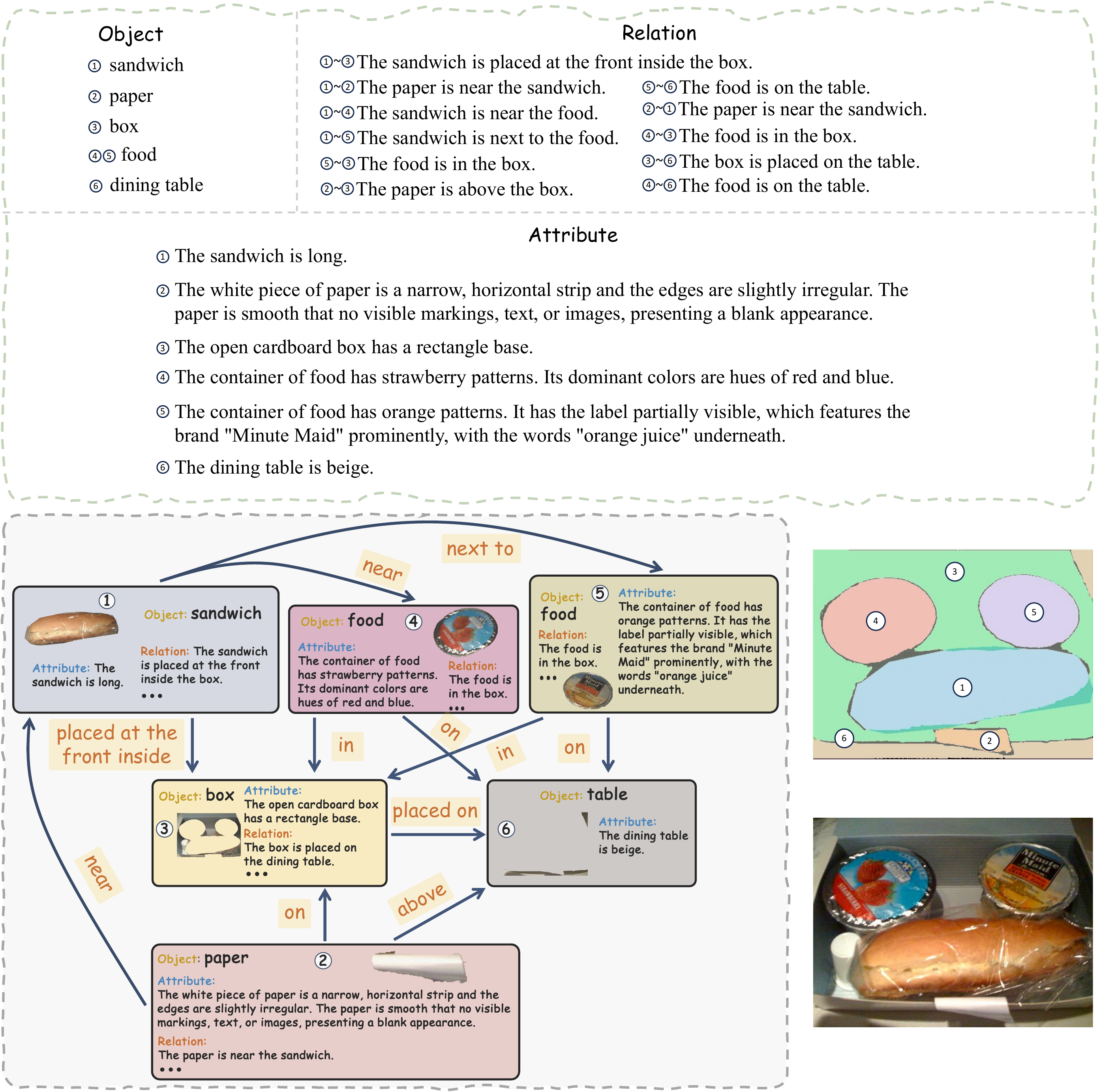}
    \caption{Data samples of \method benchmark for evaluating comprehensive captions.}
    \label{supp_fig:detail_caption_eval_data_1}
    \vspace{-2mm}
\end{figure*}

\begin{figure*}
    \centering
    \includegraphics[width=0.99\textwidth]{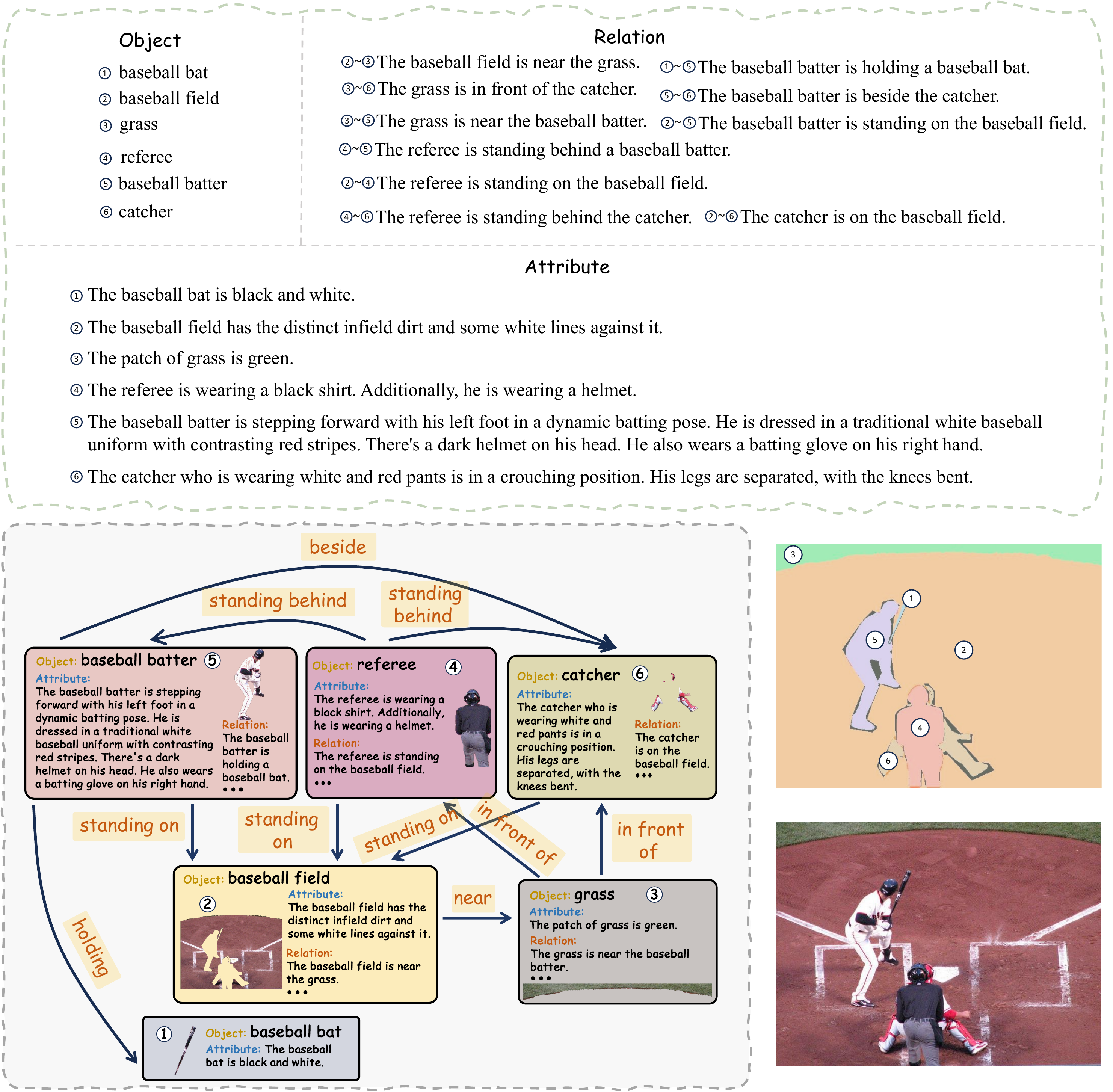}
    \caption{Data samples of \method benchmark for evaluating comprehensive captions.}
    \label{supp_fig:detail_caption_eval_data_2}
    \vspace{-2mm}
\end{figure*}

\begin{figure*}
    \centering
    \includegraphics[width=0.95\textwidth]{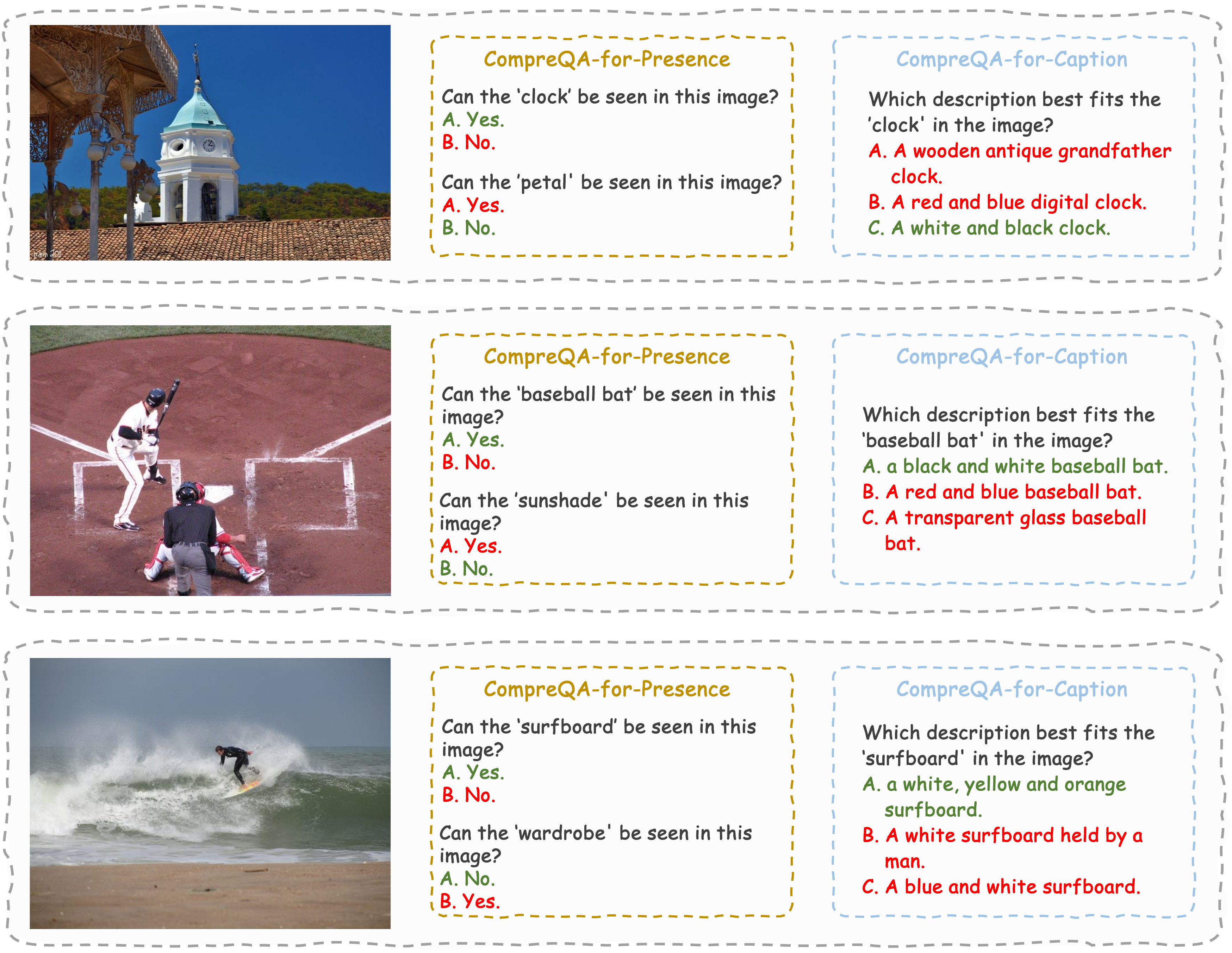}
    \vspace{-2mm}
    \caption{Data samples of \method benchmark for fine-grained object VQA. The answer is denoted in green, while the error options are in red. In the CompreQA-for-Presence task, an equal number of objects that are present and absent are included.}
    \label{supp_fig:supp_case_qa}
    \vspace{-2mm}
\end{figure*}

We present more annotation examples of comprehensive caption evaluation and fine-grained objects VQA from \method in~\cref{supp_fig:detail_caption_eval_data_1}, \cref{supp_fig:detail_caption_eval_data_2}, and~\cref{supp_fig:supp_case_qa}. The structure of directed scene graph illustrated in~\cref{supp_fig:detail_caption_eval_data_1} and~\cref{supp_fig:detail_caption_eval_data_2} is composed of the human annotation at the levels of object, attribute, and relation.
As shown in~\cref{supp_fig:supp_case_qa}, the CompreQA-for-Presence dataset includes an equal number of objects that are present and absent, and CompreQA-for-Caption dataset contains one accurate description of an object and two inaccurate descriptions.

\section{Discussion about More Long Caption Datasets}
Considering there are several recent works on dense caption datasets such as DOCCI~\cite{onoe2024docci}, ImageInWords(IIW)~\cite{garg2024imageinwords} and DCI~\cite{DCI}, we show the comparison between these datasets and our \method in~\cref{supp_tab:data_composition}. The IIW~\cite{garg2024imageinwords} and DOCCI~\cite{onoe2024docci} datasets do not provide special annotations for objects, attributes, relations, and segmentation maps. Each image within the two datasets is only annotated with one long text. Moreover, DCI~\cite{DCI} dataset provides semantic labels, segmentation maps and attribute descriptions for each object, but it doesn't consider the relationships between objects. Thus, the datasets in these works can not be utilized to comprehensively evaluate detailed captions on multiple levels as ours.

\section{Information of Human Participants}
In the construction and experimentation process of \method, three stages required the involvement of human experts: data annotation, manual captioning of images for evaluating human performance, and manual scoring of captions generated by models and humans. We engaged a total of 50 participants from universities and a crowdsourcing platform, dividing them into groups of 20, 10, and 20 participants to respectively take part in these three stages, ensuring no overlap of personnel across the stages. These participants are aged 22 to 45, with backgrounds in linguistics, computer science, and mathematics.

During the manual captioning of images, the 10 human experts each described 56 different images as comprehensively as possible. They were not specifically instructed to focus on objects, attributes, or relations in their descriptions. During the manual evaluation scoring stage, the 20 experts independently scored all 6,160 captions generated by both models and humans according to their personal assessment criteria. The average score was used as the final score for both the models and human performance.

\section{Limitations and Broader Impact}
\noindent\textbf{Broader Impact.}  We propose a human-annotated \method benchmark, which is composed in the format of directed scene graph, for evaluating comprehensive captions generated by LVLMs. Using the \method benchmark, we identify which LVLM is better at accurately describing text-rich visual content. Additionally, we design a vision question answering (VQA) task based on tiny objects to assess the fine-grained object perception ability of LVLMs.

\noindent\textbf{Limitations.} We have discussed that the qualities of detailed captions, including scores at the levels of object, attributes and relations, are not correlated to caption length generated by LVLMs. However, we do not quantitatively evaluate hallucinations (\ie, incorrect descriptions and non-existent visual information) generated by LVLMs. We plan to assess hallucination components with error rates in future work.

\end{document}